# Finding Local Minima via Stochastic Nested Variance Reduction

Dongruo Zhou* and Pan Xu† and Quanquan Gu‡

## Abstract

We propose two algorithms that can find local minima faster than the state-of-the-art algorithms in both finite-sum and general stochastic nonconvex optimization. At the core of the proposed algorithms is One-epoch-SNVRG$^+$ using stochastic nested variance reduction (Zhou et al., 2018a), which outperforms the state-of-the-art variance reduction algorithms such as SCSG (Lei et al., 2017). In particular, for finite-sum optimization problems, the proposed SNVRG$^+$ + Neon2$^{\text{finite}}$ algorithm achieves $\widetilde{O}(n^{1/2}\epsilon^{-2} + n\epsilon_H^{-3} + n^{3/4}\epsilon_H^{-7/2})$[1] gradient complexity[2] to converge to an $(\epsilon, \epsilon_H)$-second-order stationary point, which outperforms SVRG + Neon2$^{\text{finite}}$ (Allen-Zhu and Li, 2017), the best existing algorithm, in a wide regime. For general stochastic optimization problems, the proposed SNVRG$^+$ + Neon2$^{\text{online}}$ achieves $\widetilde{O}(\epsilon^{-3} + \epsilon_H^{-5} + \epsilon^{-2}\epsilon_H^{-3})$ gradient complexity, which is better than both SVRG + Neon2$^{\text{online}}$ (Allen-Zhu and Li, 2017) and Natasha2 (Allen-Zhu, 2017) in certain regimes. Furthermore, we explore the acceleration brought by third-order smoothness of the objective function.

## 1 Introduction

We study the following nonconvex optimization problem

$$\min_{\mathbf{x} \in \mathbb{R}^d} F(\mathbf{x}), \tag{1.1}$$

where $F$ is a nonconvex smooth function. Such nonconvex optimization lies at the core of various machine learning applications due to the ubiquitousness of nonconvexity. However, finding the global minimum of a nonconvex optimization problem can be generally NP hard (Hillar and Lim, 2013). Therefore, instead of finding the global minimum, various optimization methods have been developed to find an $\epsilon$-approximate first-order stationary point of (1.1), i.e., a point $\mathbf{x}$ satisfying $\|\nabla F(\mathbf{x})\|_2 \leq \epsilon$, where $\epsilon > 0$ is a predefined precision parameter. This vast body of literature consists of gradient descent (GD), stochastic gradient descent (SGD), stochastic variance reduced gradient (SVRG) (Reddi et al., 2016; Allen-Zhu and Hazan, 2016), stochastically controlled stochastic gradient (SCSG) (Lei et al., 2017) and stochastic nested variance-reduced gradient (SNVRG) (Zhou et al., 2018a).

---

*Department of Computer Science, University of Virginia, Charlottesville, VA 22904, USA; e-mail: dz4bd@virginia.edu

†Department of Computer Science, University of Virginia, Charlottesville, VA 22904, USA; e-mail: px3ds@virginia.edu

‡Department of Computer Science, University of Virginia, Charlottesville, VA 22904, USA; e-mail: qg5w@virginia.edu

[1]$\widetilde{O}(\cdot)$ hides constant and logarithmic factors.

[2]In this paper, we will use gradient complexity as a measure of an algorithm's running time, which is defined to be the total number of stochastic gradient evaluations to achieve an $(\epsilon, \epsilon_H)$-approximate local minimum.

Due to the nonconvexity of the objective function in (1.1), first-order stationary points are not always satisfying since they can be saddle points and even local maxima. To avoid such unsatisfactory stationary points, one can further pursue an $(\epsilon, \epsilon_H)$-approximate second-order stationary point (Nesterov and Polyak, 2006) of (1.1), namely a point $\mathbf{x}$ that satisfies

$$\|\nabla F(\mathbf{x})\|_2 \leq \epsilon, \text{ and } \lambda_{\min}\big(\nabla^2 F(\mathbf{x})\big) \geq -\epsilon_H, \quad (1.2)$$

where $\epsilon, \epsilon_H \in (0, 1)$ are predefined precision parameters and $\lambda_{\min}(\cdot)$ denotes the minimum eigenvalue of a matrix. An $(\epsilon, \sqrt{\epsilon})$-approximate second-order stationary point is considered as an approximate local minimum of (1.1) (Nesterov and Polyak, 2006). In many tasks such as training a deep neural network, matrix completion and matrix sensing, one have found that local minima have a very good generalization performance (Choromanska et al., 2015; Dauphin et al., 2014) or all local minima are global minima (Ge et al., 2016; Bhojanapalli et al., 2016; Zhang et al., 2018).

One particular class of algorithms to find the second-order stationary point is the cubic regularized Newton's method (Nesterov and Polyak, 2006) and all its variants (Agarwal et al., 2016; Carmon and Duchi, 2016; Curtis et al., 2017; Kohler and Lucchi, 2017; Xu et al., 2017; Tripuraneni et al., 2017; Zhou et al., 2018b), which use the Hessian information of the objective function at each iteration. Some recent work proposes to avoid the computation of the Hessian matrix in cubic regularization by approximate matrix inversion (Agarwal et al., 2016) or by accelerated gradient descent (Carmon and Duchi, 2016). Another class of algorithms has focused on using negative curvature direction to find the local minimum (Carmon et al., 2016; Allen-Zhu, 2017). However, the computation of a negative curvature direction is expensive when the optimization problem is of high dimension. Recently, there has emerged a large body of work (Xu and Yang, 2017; Allen-Zhu and Li, 2017; Jin et al., 2017b; Daneshmand et al., 2018) that only use first-order oracles to find the negative curvature direction. Specifically, Xu and Yang (2017) proposed a negative curvature originated from noise (NEON) algorithm that can extract the negative curvature direction based on gradient evaluation, which saves Hessian-vector computation. Later, Allen-Zhu and Li (2017) proposed a Neon2 algorithm, which further reduces the number of (stochastic) gradient evaluations required by NEON. Equipped with NEON and Neon2, many aforementioned algorithms such as GD, SGD, SVRG, SCSG for finding the first-order stationary point can be turned into local minimum finding ones (Xu and Yang, 2017; Allen-Zhu and Li, 2017; Yu et al., 2017b,a).

In this paper, following our recent work on stochastic nested variance reduction (Zhou et al., 2018a) for finding the first-order stationary point in nonconvex optimization, we take a step further along this line to find the second-order stationary point. More specifically, we present two novel algorithms that can find local minima faster than existing algorithms (Xu and Yang, 2017; Allen-Zhu and Li, 2017; Yu et al., 2017a) in a wide regime for both finite-sum and stochastic optimization. At the core of the proposed local minimum finding algorithms is a variant of One-epoch-SNVRG proposed in Zhou et al. (2018a) to find a first-order stationary point, namely One-epoch-SNVRG$^+$. In particular, One-epoch-SNVRG$^+$ randomly chooses the total number of iterations according to a geometric distribution, which is in a way similar to SCSG (Lei et al., 2017). It therefore enjoys a convergence guarantee regarding the last iterate of the epoch rather than a uniformly sampled iterate of the epoch. The proposed algorithms essentially use Neon2 (Allen-Zhu and Li, 2017) to turn One-epoch-SNVRG$^+$ into a local minimum finder. Here we highlight the major contributions of this paper as follows:

- We present a new variant of One-epoch-SNVRG (Zhou et al., 2018a), called One-epoch-SNVRG$^+$, which is based on stochastic nested variance reduction. Different from One-epoch-SNVRG that uses a deterministic epoch length and randomly chooses an iterate as its output, the proposed One-epoch-SNVRG$^+$ algorithm uses a random epoch length but chooses the last iterate of the epoch as its output. We derive a theoretical guarantee for One-epoch-SNVRG$^+$ with respect to the last iterate of the epoch, which enables us to integrate One-epoch-SNVRG$^+$ with Neon2 (Allen-Zhu and Li, 2017) to design a fast local minimum finding algorithm.

- In the finite-sum optimization setting, (1.1) can be written as follows

$$\min_{\mathbf{x} \in \mathbb{R}^d} F(\mathbf{x}) = \frac{1}{n} \sum_{i=1}^n f_i(\mathbf{x}), \qquad (1.3)$$

  where each $f_i$ is nonconvex, and $n$ is the number of component functions. The proposed algorithm, SNVRG$^+$ + Neon2$^{\text{finite}}$, can find an $(\epsilon, \epsilon_H)$ second-order stationary point of (1.3) within $\widetilde{O}(n^{1/2}\epsilon^{-2} + n\epsilon_H^{-3} + n^{3/4}\epsilon_H^{-7/2})$ stochastic gradient evaluations, which is evidently faster than the best existing algorithm SVRG + Neon2$^{\text{finite}}$ (Allen-Zhu and Li, 2017) that attains $\widetilde{O}(n^{2/3}\epsilon^{-2} + n\epsilon_H^{-3} + n^{3/4}\epsilon_H^{-7/2})$ gradient complexity in a wide regime. A thorough comparison is illustrated in Figure 1.

- In the general stochastic optimization setting, (1.1) is in the following form

$$\min_{\mathbf{x} \in \mathbb{R}^d} F(\mathbf{x}) = \mathbb{E}_{\xi \sim \mathcal{D}}[F(\mathbf{x}; \xi)], \qquad (1.4)$$

  where $\xi$ is a random variable drawn from some fixed but unknown distribution $\mathcal{D}$ and $F(\mathbf{x}; \xi)$ is nonconvex. Our proposed algorithm, SNVRG$^+$ + Neon2$^{\text{online}}$, can find an $(\epsilon, \epsilon_H)$ second-order stationary point of (1.4) within $\widetilde{O}(\epsilon^{-3} + \epsilon_H^{-5} + \epsilon^{-2}\epsilon_H^{-3})$ stochastic gradient evaluations, which is again faster than the state-of-the-art algorithms such as SCSG+Neon2$^{\text{online}}$ (Allen-Zhu and Li, 2017) with $\widetilde{O}(\epsilon^{-10/3} + \epsilon_H^{-5} + \epsilon^{-2}\epsilon_H^{-3})$ gradient complexity, and Natasha2 (Allen-Zhu, 2017) with $\widetilde{O}(\epsilon^{-3.25} + \epsilon^{-3}\epsilon_H + \epsilon_H^{-5})$ gradient complexity in certain regime. A detailed comparison is demonstrated in Figure 2.

- We also show that our proposed algorithms can find local minima even faster when the objective function enjoys the third-order smoothness property. We prove that our proposed algorithms achieve faster convergence rates to a local minimum than the FLASH algorithm proposed in Yu et al. (2017a), which also exploits the third-order smoothness of objective functions for both finite-sum and general stochastic optimization problems.

The remainder of this paper is organized as follows: In Section 2 we review the relevant work in the literature. We present preliminary definitions in Section 3. We then present our proposed algorithms in Section 4. We present our main theoretical results for second-order smooth functions in Section 5 and that for third-order smooth functions in Section 6. We conclude the paper with Section 7.

**Notation:** Denote $\mathbf{A} = [A_{ij}] \in \mathbb{R}^{d \times d}$ as a matrix and $\mathbf{x} = (x_1, ..., x_d)^\top \in \mathbb{R}^d$ as a vector. $\|\mathbf{v}\|_2$ denotes the 2-norm of a vector $\mathbf{v} \in \mathbb{R}^d$. We use $\langle \cdot, \cdot \rangle$ to represent the inner product. For two sequences $\{a_n\}$ and $\{b_n\}$, we denote $a_n = O(b_n)$ if there is a constant $0 < C < +\infty$ such that $a_n \leq C b_n$, denote $a_n = \Omega(b_n)$ if there is a constant $0 < C < +\infty$, such that $a_n \geq C b_n$, and use $\widetilde{O}(\cdot)$ to hide logarithmic factors. We also write $a_n \lesssim b_n$ (or $a_n \gtrsim b_n$) if $a_n$ is less than (or larger than) $b_n$ up to a constant. We denote the product $c_a c_{a+1} \ldots c_b$ term as $\prod_{i=a}^{b} c_i$. In addition, if $a > b$, we define $\prod_{i=a}^{b} c_i = 1$. In this paper, $\lfloor \cdot \rfloor$ represents the floor function and $\log(x)$ represents the logarithm of $x$ to base 2. $a \wedge b$ means $\min(a, b)$. We denote by $\mathbb{1}\{\mathcal{E}\}$ the indicator function such that $\mathbb{1}\{\mathcal{E}\} = 1$ if the event $\mathcal{E}$ is true, and $\mathbb{1}\{\mathcal{E}\} = 0$ otherwise.

## 2 Related Work

In this section, we review and discuss the relevant work. We start with algorithms for finding first-order stationary points. To find an $\epsilon$-approximate first-order stationary point, stochastic gradient descent (SGD) (Robbins and Monro, 1951) (Nesterov, 2013) and its variants (Ghadimi and Lan, 2013; Ghadimi et al., 2016) achieve $O(1/\epsilon^4)$ gradient complexity. Variance reduced techniques such as nonconvex SVRG inspired from the convex SVRG (Johnson and Zhang, 2013) have been proposed to accelerate SGD in finding the first-order stationary point (Reddi et al., 2016; Allen-Zhu and Hazan, 2016). Lei et al. (2017) proposed a new variance reduction algorithm, i.e., the stochastically controlled stochastic gradient (SCSG) algorithm, which finds a first-order stationary point within $O(\min\{\epsilon^{-10/3}, n^{2/3}\epsilon^{-2}\})$ stochastic gradient evaluations for finite-sum optimization in (1.3), and outperforms SVRG when the number of component functions $n$ is large. Very recently, Zhou et al. (2018a) proposed a stochastic nested variance reduced gradient (SNVRG) algorithm, which improves the aforementioned variance reduction methods via employing multiple reference points and gradients in the update. They proved that SNVRG enjoys $\widetilde{O}(n \wedge \epsilon^{-2} + n^{1/2}\epsilon^{-2} \wedge \epsilon^{-3})$ gradient complexity for finding an $\epsilon$-approximate first-order stationary point, and outperforms SVRG and SCSG uniformly.

The literature of finding local minima in nonconvex optimization can be roughly divided into three categories according to the oracles they use: Hessian-based, Hessian-vector product-based and gradient-based (Heesian-free). We review each category in the sequel accordingly. The most popular algorithm using Hessian matrix to find an $(\epsilon, \sqrt{\epsilon})$-approximate local minimum is the cubic regularized Newton's method (Nesterov and Polyak, 2006), which attains $O(\epsilon^{-3/2})$ iteration complexity. The trust region method is proved to achieve the same iteration complexity (Curtis et al., 2017). To alleviate the computation burden of evaluating full gradients and Hessian matrices in large-scale optimization problems, subsampled cubic regularization and trust-region methods (Kohler and Lucchi, 2017; Xu et al., 2017) were proposed and proved to enjoy the same iteration complexity as their original versions with full gradients and Hessian matrices. Recently, stochastic variance reduced cubic regularization method (SVRC) (Zhou et al., 2018b) was proposed, which achieves the best-known second-order oracle complexity among existing cubic regularization methods.

Another line of research uses Hessian-vector products to find the second-order stationary points. Carmon et al. (2016) Agarwal et al. (2016) independently proposed two algorithms that can find an $(\epsilon, \sqrt{\epsilon})$-approximate local minimum within $O(\epsilon^{-7/4})$ full gradient and Hessian-product evaluations. Agarwal et al. (2016) also showed that their algorithm only needs $O(n\epsilon^{-3/2} + n^{3/4}\epsilon^{7/4})$ stochastic gradient and Hessian-vector product evaluations for finite-sum optimization problems (1.3). Reddi et al. (2017) proposed a generic algorithmic framework that uses both first-order and second-order methods to find the local minimum within $O(n^{2/3}\epsilon^{-2} + n\epsilon^{-3/2} + n^{3/4}\epsilon^{7/4})$ stochastic gradient and Hessian-product evaluations. Allen-Zhu (2017) proposed the Natasha2 algorithm which finds an $(\epsilon, \sqrt{\epsilon})$-approximate second-order stationary point within $O(\epsilon^{-7/2})$ stochastic gradient and Hessian-vector product evaluations.

The last line of research uses purely gradient information to find the local minima. Ge et al. (2015); Levy (2016) studied the perturbed GD and SGD algorithms for escaping saddle points, where isotropic noise is added into the gradient or stochastic gradient at each iteration or whenever the gradient is sufficiently small. Jin et al. (2017a) further proposed a perturbed accelerated gradient descent, which can finds the second-order stationary point even faster. Xu and Yang (2017) showed that perturbed gradient or stochastic gradient descent can help find the negative curvature direction without using Hessian matrix and proposed the NEON algorithm that extracts the negative curvature using only first-order information. Later Allen-Zhu and Li (2017) developed the Neon2 algorithm, which improves upon on Neon, and turns Natasha2 (Allen-Zhu, 2017) into a first-order method to find the local minima. Yu et al. (2017b) proposed the gradient descent with one-step escaping algorithm (GOSE) that saves negative curvature computation and Yu et al. (2017a) proposed the FLASH algorithm that exploits the third-order smoothness of the objective function. Very recently, Daneshmand et al. (2018) proved that SGD with periodically changing step size can escape from saddle points under an additional correlated negative curvature (CNC) assumption on the stochastic gradient.

To give a comprehensive comparison of different algorithms for finding local minima, we present the gradient complexities of all aforementioned first-order local minimum finding algorithms in Table 1. We can see from Table 1 that our proposed algorithms SNVRG$^+$ + Neon2$^{\text{finite}}$ and SNVRG$^+$ + Neon2$^{\text{online}}$ outperform all other first-order algorithms in finding an $(\epsilon, \epsilon_H)$-approximate second-order stationary point for nonconvex optimization problems in a wide regime, for both finite-sum and general stochastic optimization.

## 3 Preliminaries

In this section, we provide the basic definitions that will be used throughout this paper.

**Definition 3.1** (Smoothness). $f : \mathbb{R}^d \to \mathbb{R}$ is $L_1$-smooth for some constant $L_1 > 0$, if it is differentiable and satisfies

$$\|\nabla f(\mathbf{x}) - \nabla f(\mathbf{y})\|_2 \leq L_1 \|\mathbf{x} - \mathbf{y}\|_2, \quad \text{for any } \mathbf{x}, \mathbf{y} \in \mathbb{R}^d.$$

**Definition 3.2** (Hessian Lipschitz). $f : \mathbb{R}^d \to \mathbb{R}$ is $L_2$-Hessian Lipschitz for some constant $L_2 > 0$, if it is twice-differentiable and satisfies

$$\|\nabla^2 f(\mathbf{x}) - \nabla^2 f(\mathbf{y})\|_2 \leq L_2 \|\mathbf{x} - \mathbf{y}\|_2, \quad \text{for any } \mathbf{x}, \mathbf{y} \in \mathbb{R}^d.$$

The above two smoothness conditions are widely used in nonconvex optimization problems (Nesterov and Polyak, 2006). We will call them first-order smoothness and second-order smoothness respectively in this paper. As shown in Carmon et al. (2017); Yu et al. (2017a), when the objective function has additionally third-order smoothness, one can design algorithms that find local minima even faster. Following Yu et al. (2017a), we denote the three-way tensor $\nabla^3 f(\mathbf{x}) \in \mathbb{R}^{d \times d \times d}$ as the third-order derivative of $f$.

**Definition 3.3** (Third-order Derivative). The third-order derivative of function $f : \mathbb{R}^d \to \mathbb{R}$ is defined as a three-way tensor $\nabla^3 f(\mathbf{x}) \in \mathbb{R}^{d \times d \times d}$, where

$$[\nabla^3 f(\mathbf{x})]_{ijk} = \frac{\partial}{\partial x_i \partial x_j \partial x_k} f(\mathbf{x}), \quad i, j, k = 1, \ldots, d \text{ and } \mathbf{x} \in \mathbb{R}^d.$$

Table 1: Comparisons on gradient complexities of different algorithms to find an $(\epsilon, \epsilon_H)$-approximate second-order stationary point in both finite-sum and general stochastic optimization settings. The last column indicates whether the algorithm exploits the third-order smoothness of the objective function.

| Setting | Algorithm | Gradient complexity | 3rd-order smooth |
|---|---|---|---|
| Finite-Sum | PGD (Jin et al., 2017a) | $\widetilde{O}\left(\frac{n}{\epsilon^2}\right)$ (for $\epsilon_H \geq \epsilon^{1/2}$) | No |
| | SVRG + Neon2$^{\text{finite}}$ (Allen-Zhu and Li, 2017) | $\widetilde{O}\left(\frac{n^{2/3}}{\epsilon^2} + \frac{n}{\epsilon_H^3} + \frac{n^{3/4}}{\epsilon_H^{7/2}}\right)$ | No |
| | FLASH (Yu et al., 2017a) | $\widetilde{O}\left(\frac{n^{2/3}}{\epsilon^2} + \frac{n}{\epsilon_H^2} + \frac{n^{3/4}}{\epsilon_H^{5/2}}\right)$ | Needed |
| | SNVRG$^+$ + Neon2$^{\text{finite}}$ (Algorithm 2) | $\widetilde{O}\left(\frac{n^{1/2}}{\epsilon^2} + \frac{n}{\epsilon_H^3} + \frac{n^{3/4}}{\epsilon_H^{7/2}}\right)$ | No |
| | SNVRG$^+$ + Neon2$^{\text{finite}}$ (Algorithm 2) | $\widetilde{O}\left(\frac{n^{1/2}}{\epsilon^2} + \frac{n}{\epsilon_H^2} + \frac{n^{3/4}}{\epsilon_H^{5/2}}\right)$ | Needed |
| Stochastic | Perturbed SGD (Ge et al., 2015) | $\widetilde{O}\left(\frac{\text{poly}(d)}{\epsilon^4}\right)$ (for $\epsilon_H \geq \epsilon^{1/4}$) | No |
| | CNC-SGD (Daneshmand et al., 2018) | $\widetilde{O}\left(\frac{1}{\epsilon^4} + \frac{1}{\epsilon_H^{10}}\right)$ | No |
| | Natasha2+Neon2$^{\text{online}}$ (Allen-Zhu, 2017) | $\widetilde{O}\left(\frac{1}{\epsilon^{3.25}} + \frac{1}{\epsilon^3 \epsilon_H} + \frac{1}{\epsilon_H^5}\right)$ | No |
| | SCSG+Neon2$^{\text{online}}$ (Allen-Zhu and Li, 2017) | $\widetilde{O}\left(\frac{1}{\epsilon^{10/3}} + \frac{1}{\epsilon^2 \epsilon_H^3} + \frac{1}{\epsilon_H^5}\right)$ | No |
| | FLASH (Yu et al., 2017a) | $\widetilde{O}\left(\frac{1}{\epsilon^{10/3}} + \frac{1}{\epsilon^2 \epsilon_H^2} + \frac{1}{\epsilon_H^4}\right)$ | Needed |
| | SNVRG$^+$ + Neon2$^{\text{online}}$ (Algorithm 3) | $\widetilde{O}\left(\frac{1}{\epsilon^3} + \frac{1}{\epsilon^2 \epsilon_H^3} + \frac{1}{\epsilon_H^5}\right)$ | No |
| | SNVRG$^+$ + Neon2$^{\text{online}}$ (Algorithm 3) | $\widetilde{O}\left(\frac{1}{\epsilon^3} + \frac{1}{\epsilon^2 \epsilon_H^2} + \frac{1}{\epsilon_H^4}\right)$ | Needed |

Now we are ready to present the formal definition of third-order smoothness, which has been explored in Anandkumar and Ge (2016); Carmon et al. (2017); Yu et al. (2017a). It is also called third-order derivative Lipschitzness in Carmon et al. (2017).

**Definition 3.4** (Third-order smoothness). $f : \mathbb{R}^d \to \mathbb{R}$ is $L_3$-third-order smooth for some constant $L_3 > 0$, if it is thrice-differentiable and satisfies
$$\|\nabla^3 f(\mathbf{x}) - \nabla^3 f(\mathbf{y})\|_F \leq L_3 \|\mathbf{x} - \mathbf{y}\|_2, \quad \text{for any } \mathbf{x}, \mathbf{y} \in \mathbb{R}^d.$$

The following definition characterizes the distance between the initial point of an algorithm and the minimizer of function $f$.

**Definition 3.5** (Optimal Gap). The optimal gap of $f$ at point $\mathbf{x}_0$ is denoted by $\Delta_f$ and
$$f(\mathbf{x}_0) - \inf_{\mathbf{x} \in \mathbb{R}^d} f(\mathbf{x}) \leq \Delta_f.$$

W.L.O.G., we assume $\Delta_f < +\infty$.

Inspired by the SCSG algorithm (Lei et al., 2017), we will use the property of geometric distribution in our algorithm design. The definition of geometric random variable is as follows.

**Definition 3.6** (Geometric Distribution). A random variable $X$ follows a geometric distribution with parameter $p$, denoted as $\text{Geom}(p)$, if it holds that

$$\mathbb{P}(X = k) = p(1-p)^k, \quad \forall k = 0, 1, \ldots.$$

**Definition 3.7** (Sub-Gaussian Stochastic Gradient). We say a function $F$ has $\sigma^2$-sub-Gaussian stochastic gradient $\nabla F(\mathbf{x}; \xi)$ for any $\mathbf{x} \in \mathbb{R}^d$ and random variable $\xi \sim \mathcal{D}$, if it satisfies

$$\mathbb{E}\bigg[\exp\bigg(\frac{\|\nabla F(\mathbf{x}; \xi) - \nabla f(\mathbf{x})\|_2^2}{\sigma^2}\bigg)\bigg] \leq \exp(1).$$

Note that Definition 3.7 implies $\mathbb{E}[\|\nabla F(\mathbf{x}; \xi) - \nabla f(\mathbf{x})\|_2^2] \leq 2\sigma^2$ (Vershynin, 2010). In the finite-sum optimization setting (1.3), we call $\nabla f_i(\mathbf{x})$ a stochastic gradient of function $F$ for a randomly chosen index $i \in [n]$, and we say $F$ has $\sigma^2$-sub-Gaussian stochastic gradient if $\mathbb{E}[\|\nabla f_i(\mathbf{x}) - \nabla F(\mathbf{x})\|_2^2] \leq 2\sigma^2$.

## 4 The Proposed Algorithms

In this section, we present our main algorithms that are built upon a variant of one-epoch stochastic nested variance reduced gradient (One-epoch-SNVRG) (Zhou et al., 2018a) and Neon2 (Allen-Zhu and Li, 2017) to find a local minimum in nonconvex optimization faster than existing methods.

### 4.1 One-epoch-SNVRG$^+$: Finding First-order Stationary Points

We first propose One-epoch-SNVRG$^+$ algorithm in Algorithm 1, which helps find a first-order stationary point faster than existing algorithms such as SCSG (Lei et al., 2017). At the core of Algorithm 1 is the idea of using multiple reference points and reference gradients to accelerate the original SVRG algorithm (Reddi et al., 2016; Allen-Zhu and Hazan, 2016), which is inspired from Zhou et al. (2018a). Note that our Algorithm 1 is different from the One-epoch-SNVRG algorithm in Zhou et al. (2018a) regarding the choice of the number of iteration $T$ which is chosen to be a random variable following a geometric distribution rather than fixed. Another difference is that in Algorithm 1, it outputs the last iterate $\mathbf{x}_T$ (Line 10) instead of uniformly choosing the output from $\mathbf{x}_1, \ldots, \mathbf{x}_T$ as in Zhou et al. (2018a). We will show in the next section that these differences are essential in the theoretical analysis of finding local minima.

### 4.2 SNVRG$^+$ + Neon2: Finding Local Minima

Based on One-epoch-SNVRG$^+$ and Neon2 (Allen-Zhu and Li, 2017), we are ready to present our main algorithms, which employ One-epoch-SNVRG$^+$ to find the first-order stationary point, and use Neon2 to escape strict saddle points and converge to a local minimum of nonconvex optimization problems. Specifically, we proposed two different algorithms for solving the finite-sum optimization problem in (1.3) and the general stochastic optimization problem in (1.4) respectively.

**Algorithm 1** One-epoch-SNVRG$^+$($\mathbf{x}_0$,$F$,$K$,$M$,$\{T_l\}$,$\{B_l\}$)

1: **Input:** initial point $\mathbf{x}_0$, function $F$, loop number $K$, step size parameter $M$, loop parameters $T_l, l \in [K]$, batch parameters $B_l, l \in [K]$, base batch size $B_0 > 0$.
2: Generate $T \sim \text{Geom}(1/(1 + \prod_{l=1}^K T_l))$
3: **for** $t = 0, ..., T - 1$ **do**
4:     $r = \min\{j : 0 = (t \mod \prod_{l=j+1}^K T_l), 0 \leq j \leq K\}$
5:     $\{\mathbf{x}_t^{(l)}\} \leftarrow \text{Update\_reference\_points}(\{\mathbf{x}_{t-1}^{(l)}\}, \mathbf{x}_t, r), 0 \leq l \leq K$.
6:     $\{\mathbf{g}_t^{(l)}\} \leftarrow \text{Update\_reference\_gradients}(\{\mathbf{g}_{t-1}^{(l)}\}, \{\mathbf{x}_t^{(l)}\}, r), 0 \leq l \leq K$.
7:     $\mathbf{v}_t \leftarrow \sum_{l=0}^K \mathbf{g}_t^{(l)}$
8:     $\mathbf{x}_{t+1} \leftarrow \mathbf{x}_t - 1/(10M) \cdot \mathbf{v}_t$
9: **end for**
10: **Output:** $\mathbf{x}_T$

---

11: **Function:** Update\_reference\_points($\{\mathbf{x}_{\text{old}}^{(l)}\}, \mathbf{x}, r$)
12: $\mathbf{x}_{\text{new}}^{(l)} \leftarrow \mathbf{x}_{\text{old}}^{(l)}, 0 \leq l \leq r-1; \mathbf{x}_{\text{new}}^{(l)} \leftarrow \mathbf{x}, r \leq l \leq K$
13: **return** $\{\mathbf{x}_{\text{new}}^{(l)}\}$

---

14: **Function:** Update\_reference\_gradients($\{\mathbf{g}_{\text{old}}^{(l)}\}, \{\mathbf{x}_{\text{new}}^{(l)}\}, r$)
15: **if** $r > 0$ **then**
16:     $\mathbf{g}_{\text{new}}^{(l)} \leftarrow \mathbf{g}_{\text{old}}^{(l)}, 0 \leq l < r; \mathbf{g}_{\text{new}}^{(l)} \leftarrow 0, r+1 \leq l \leq K$
17:     Uniformly generate index set $I \subset [n]$ without replacement, $|I| = B_r$
18:     $\mathbf{g}_{\text{new}}^{(r)} \leftarrow 1/B_r \sum_{i \in I} \left[ \nabla f_i(\mathbf{x}_{\text{new}}^{(r)}) - \nabla f_i(\mathbf{x}_{\text{new}}^{(r-1)}) \right]$
19: **else**
20:     Uniformly generate index set $I \subset [n]$ without replacement, $|I| = B_0$
21:     $\mathbf{g}_{\text{new}}^{(0)} \leftarrow 1/B_0 \sum_{i \in I} \nabla f_i(\mathbf{x}_{\text{new}}^{(0)}); \mathbf{g}_{\text{new}}^{(l)} \leftarrow 0, 1 \leq l \leq K$
22: **end if**
23: **return** $\{\mathbf{g}_{\text{new}}^{(l)}\}$.

---

To solve the finite-sum optimization problem (1.3), we propose the SNVRG$^+$ + Neon2$^{\text{finite}}$ algorithm to find the local minimum, which is displayed in Algorithm 2. At each iteration of 2, it first determines whether the current point is a first-order stationary point (Line 3) or not. If not, it will run Algorithm 1 (One-epoch-SNVRG$^+$) in order to find a first-order stationary point. Once obtaining a first-order stationary point, it will call Neon2$^{\text{finite}}$ to find the negative curvature direction to escape any potential non-degenerate saddle point. If Neon2$^{\text{finite}}$ does not find such a direction, it will output $\widehat{\mathbf{v}} = \perp$ and Algorithm 2 terminates and outputs $\mathbf{z}_{u-1}$ (Line 8) since it has already reached a second-order stationary point according to (1.2). If Neon2$^{\text{finite}}$ finds a negative curvature direction $\widehat{\mathbf{v}} \neq \perp$, Algorithm 2 will perform one step of negative curvature descent in the direction of $\widehat{\mathbf{v}}$ or $-\widehat{\mathbf{v}}$ (Line 11) to escape the non-degenerate saddle point.

To solve the general stochastic optimization problem in (1.4), we propose the SNVRG$^+$ + Neon2$^{\text{online}}$ algorithm to find the local minimum, which is displayed in Algorithm 3. It is almost the same as Algorithm 2 used in the finite-sum nonconvex optimization setting except that it uses a subsampled gradient to determine whether we have obtained a first-order stationary point (Line 4 in Algorithm 3) and it uses Neon2$^{\text{online}}$ (Line 7) to find the negative curvature direction to escape

**Algorithm 2** SNVRG$^+$ + Neon2$^{\text{finite}}$($\mathbf{z}_0, F, K, M, \{T_l\}, \{B_l\}, U, \epsilon, \epsilon_H, \delta, \eta, L_1, L_2$)

1: **for** $u = 1, \ldots, U$ **do**
2:     $\mathbf{g}_{u-1} = \nabla F(\mathbf{z}_{u-1})$
3:     **if** $\|\mathbf{g}_{u-1}\|_2 \geq \epsilon$ **then**
4:         $\mathbf{z}_u = $ One-epoch-SNVRG$^+$($\mathbf{z}_{u-1}, F, K, M, \{T_l\}, \{B_l\}$)     ▷ Algorithm 1
5:     **else**
6:         $\mathbf{v} = $ Neon2$^{\text{finite}}$($F, \mathbf{z}_{u-1}, L_1, L_2, \delta, \epsilon_H$)
7:         **if** $\mathbf{v} = \perp$ **then**
8:             **return** $\mathbf{z}_{u-1}$
9:         **else**
10:            Generate a Rademacher random variable $\zeta$
11:            $\mathbf{z}_u \leftarrow \mathbf{z}_{u-1} + \zeta \eta \widehat{\mathbf{v}}$
12:         **end if**
13:     **end if**
14: **end for**
15: **return**

the potential saddle points. Algorithm 3 will terminate and output the current iterate if no negative curvature direction is found (Line 9).

**Algorithm 3** SNVRG$^+$ + Neon2$^{\text{online}}$($\mathbf{z}_0, F, K, M, \{T_l\}, \{B_l\}, U, \epsilon, \epsilon_H, \delta, \eta, L_1, L_2$)

1: **for** $u = 1, \ldots, U$ **do**
2:     Uniformly generate index set $I \subset [n]$ without replacement, $|I| = B_0$
3:     $\mathbf{g}_{u-1} = 1/B_0 \sum_{i \in I} \nabla f_i(\mathbf{z}_{u-1})$
4:     **if** $\|\mathbf{g}_{u-1}\|_2 \geq \epsilon/2$ **then**
5:         $\mathbf{z}_u = $ One-epoch-SNVRG$^+$($\mathbf{z}_{u-1}, F, K, M, \{T_l\}, \{B_l\}$)     ▷ Algorithm 1
6:     **else**
7:         $\mathbf{v} = $ Neon2$^{\text{online}}$($F, \mathbf{z}_{u-1}, L_1, L_2, \delta, \epsilon_H$)
8:         **if** $\mathbf{v} = \perp$ **then**
9:             **return** $\mathbf{z}_{u-1}$
10:         **else**
11:            Generate a Rademacher random variable $\zeta$
12:            $\mathbf{z}_u \leftarrow \mathbf{z}_{u-1} + \zeta \eta \widehat{\mathbf{v}}$
13:         **end if**
14:     **end if**
15: **end for**
16: **return**

Note that both Algorithms 2 and 3 are only based on the gradient information of the objective function and are therefore first-order optimization algorithms. As we will show in the next two sections, our proposed algorithms push the frontier of first-order stochastic optimization algorithms for finding local minima (Xu and Yang, 2017; Allen-Zhu and Li, 2017; Allen-Zhu, 2017; Yu et al., 2017b,a).

# 5 Main Theory for Finding Local Minima

In this section, we provide the main theoretical results for the proposed algorithms. We start with the following key lemma that characterizes the function value decrease of One-epoch-SNVRG$^+$.

**Lemma 5.1.** Suppose that each $f_i$ is $L_1$-smooth and $F$ has $\sigma^2$-sub-Gaussian stochastic gradient. In Algorithm 1, suppose $B_0 \geq 2$ and the number of nested loops $K = \log \log B_0$. Choose the step size $M \geq 6L_1$. For the loop length and batch size parameters, let $T_1 = 2$, $B_1 = 6^K \cdot B_0$ and

$$T_l = 2^{2^{l-2}}, \qquad B_l = 6^{K-l+1} \cdot B_0/2^{2^{l-1}},$$

for all $2 \leq l \leq K$. Then the output of Algorithm 1 $\mathbf{x}_T$ satisfies

$$\mathbb{E}\|\nabla F(\mathbf{x}_T)\|_2^2 \leq C\bigg(\frac{M}{B_0^{1/2}} \cdot \mathbb{E}[F(\mathbf{x}_0) - F(\mathbf{x}_T)] + \frac{2\sigma^2}{B_0} \cdot \mathbb{1}\{B_0 < n\}\bigg), \tag{5.1}$$

where $C = 100$. In addition, the total number of stochastic gradient computations $\mathcal{T}$ by Algorithm 1 satisfies $\mathbb{E}\mathcal{T} \leq 20B_0 \log^3 B_0$.

**Remark 5.2.** Note that a similar lemma is proved in Zhou et al. (2018a) for One-epoch-SNVRG. We would like to emphasize one crucial difference between Lemma 5.1 and Lemma 4.1 in Zhou et al. (2018a). On the left-hand side of (5.1), the expected squared norm of the gradient is regarding the last iterate of $\mathbf{x}_T$ in Algorithm 1, while the expected squared norm of the gradient in Lemma 4.1 in Zhou et al. (2018a) is regarding $\mathbf{x}_{\text{out}}$, which is a uniformly chosen iterate from $\mathbf{x}_1, \ldots, \mathbf{x}_T$. This key difference is due to the different selection of the output between One-epoch-SNVRG$^+$ in Algorithm 1 and One-epoch-SNVRG. It ensures the ergodic-type guarantee of One-epoch-SNVRG$^+$ which is essential in the analysis of local minimum finding algorithms.

**Remark 5.3.** For simplicity, we use $\nabla f_i(\mathbf{x})$ to denote the stochastic gradient at point $\mathbf{x}$ in our One-epoch-SNVRG$^+$ algorithm (Lines 18 and 21 in Algorithm 1) and the analysis of Lemma 5.1. However, we emphasize that One-epoch-SNVRG$^+$ also works in the general stochastic optimization setting if we replace $\nabla f_i(\mathbf{x})$ with $\nabla F(\mathbf{x}; \xi_i)$ for any index $i$. And the theoretical result in Lemma 5.1 still holds.

## 5.1 Finite-Sum Optimization Problems

We start with the nonconvex finite-sum optimization problem (1.3). The following theorem provides the gradient complexity of Algorithm 2 in finding an approximate local minimum.

**Theorem 5.4.** Suppose that $F = 1/n \sum_{i=1}^n f_i$, where each $f_i$ is $L_1$-smooth and $L_2$-Hessian Lipschitz continuous. Let $0 < \epsilon, \epsilon_H < 1$, $\delta = \epsilon_H^3/(144L_2^2\Delta_F)$ and $U = 24L_2^2\Delta_F\epsilon_H^{-3} + 1800L_1\Delta_F\epsilon^{-2}n^{-1/2}$. Set $B_0 = n$, $M = 6L_1$ and all the rest parameters of One-epoch-SNVRG$^+$ as in Lemma 5.1. Choose step size $\eta = \epsilon_H/L_2$. Then with probability at least $1/4$, SNVRG$^+$ + Neon2$^{\text{finite}}$ will find an $(\epsilon, \epsilon_H)$-second-order stationary point within

$$\widetilde{O}\bigg(\frac{\Delta_F n L_2^2}{\epsilon_H^3} + \frac{\Delta_F n^{3/4} L_1^{1/2} L_2^2}{\epsilon_H^{7/2}} + \frac{\Delta_F n^{1/2} L_1}{\epsilon^2}\bigg) \tag{5.2}$$

stochastic gradient evaluations.

**Remark 5.5.** Note that the gradient complexity in Theorem 5.4 holds with constant probability $1/4$. In practice, we can repeatedly run Algorithm 2 for $\log(1/p)$ times to achieve a result that holds with probability at least $1 - p$ for any $p \in (0, 1)$. Similar boosting techniques have also been used in Allen-Zhu and Li (2017); Yu et al. (2017b,a).

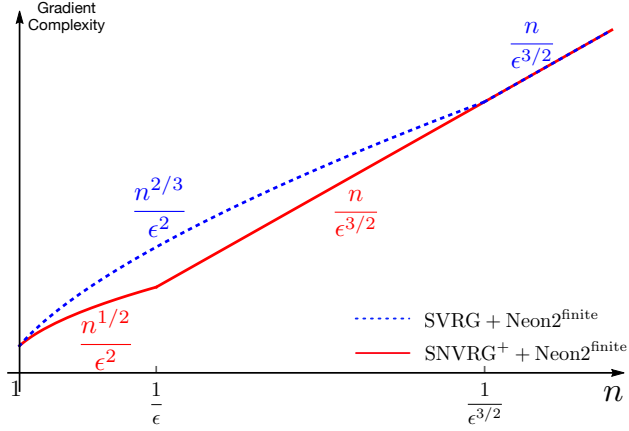

Figure 1: Comparison of gradient complexities between $\text{SNVRG}^+ + \text{Neon2}^{\text{finite}}$ and $\text{SVRG} + \text{Neon2}^{\text{finite}}$ for finding an $(\epsilon, \sqrt{\epsilon})$-approximate local minimum in finite-sum optimization problems.

**Remark 5.6.** For finite-sum nonconvex optimization, Theorem 5.4 suggests that the gradient complexity of Algorithm 2 ($\text{SNVRG}^+ + \text{Neon2}^{\text{finite}}$) is $\widetilde{O}(n^{1/2}\epsilon^{-2} + n\epsilon_H^{-3} + n^{3/4}\epsilon_H^{-7/2})$. In contrast, the gradient complexity of other state-of-the-art local minimum finding algorithms ($\text{SVRG} + \text{Neon2}^{\text{finite}}$) (Allen-Zhu and Li, 2017) is $\widetilde{O}(n^{2/3}\epsilon^{-2} + n\epsilon_H^{-3} + n^{3/4}\epsilon_H^{-7/2})$. Our algorithm is strictly better than that of Allen-Zhu and Li (2017) in terms of the first term in the big O notation.

If we choose $\epsilon_H = \sqrt{\epsilon}$, the gradient complexity of our algorithm to find an $(\epsilon, \sqrt{\epsilon})$-approximate local minimum turns out to be $O(n^{1/2}\epsilon^{-2} + n\epsilon^{-3/2} + n^{3/4}\epsilon^{-7/4})$ and that of $\text{SVRG} + \text{Neon2}^{\text{finite}}$ is $O(n^{2/3}\epsilon^{-2} + n\epsilon^{-3/2} + n^{3/4}\epsilon^{-7/4})$. We compare these two algorithms in Figure 1 when $\epsilon_H = \sqrt{\epsilon}$ and make the following comments:

- When $n \gtrsim \epsilon^{-3/2}$, the gradient complexities of both algorithms are in the same order of $\widetilde{O}(n\epsilon^{-3/2})$.

- When $\epsilon^{-1} \lesssim n \lesssim \epsilon^{-3/2}$, $\text{SNVRG}^+ + \text{Neon2}^{\text{finite}}$ enjoys $\widetilde{O}(n\epsilon^{-3/2})$ gradient complexity, which is strictly better than that of $\text{SVRG} + \text{Neon2}^{\text{finite}}$, i.e., $\widetilde{O}(n^{2/3}\epsilon^{-2})$.

- Lastly, when $n \lesssim \epsilon^{-1}$, $\text{SNVRG}^+ + \text{Neon2}^{\text{finite}}$ achieves $\widetilde{O}(n^{1/2}\epsilon^{-2})$ gradient complexity, which is again better than the gradient complexity of $\text{SVRG} + \text{Neon2}^{\text{finite}}$, $\widetilde{O}(n^{2/3}\epsilon^{-2})$, by a factor of $\widetilde{O}(n^{1/6})$.

In short, our algorithms beats $\text{SVRG} + \text{Neon2}^{\text{finite}}$ when $n \lesssim \epsilon^{-3/2}$.

### 5.2 General Stochastic Optimization Problems

Now we consider the general stochastic optimization problem (1.4). Recall that in this setting we will call $\nabla F(\mathbf{x}; \xi_i)$ the stochastic gradient at point $\mathbf{x}$ for some random variable $\xi_i$ and index $i$. In the following theorem, we present the gradient complexity of Algorithm 3 for finding local minima.

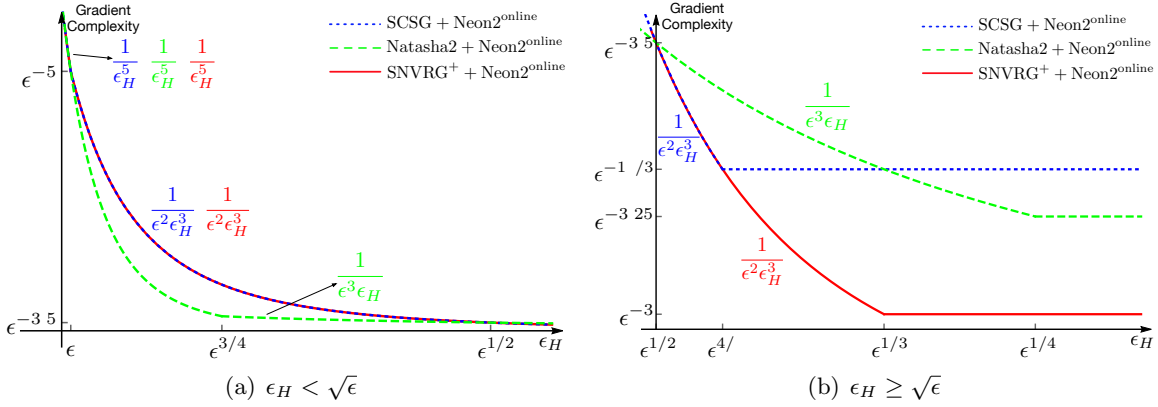

Figure 2: Comparison of gradient complexities among SNVRG$^+$ + Neon2$^{\text{online}}$, SCSG + Neon2$^{\text{online}}$ and Natasha2 + Neon2$^{\text{online}}$ for finding an $(\epsilon, \epsilon_H)$-approximate second-order stationary point: (a) the comparison when $\epsilon_H < \sqrt{\epsilon}$, and (b) the comparison when $\epsilon_H \geq \sqrt{\epsilon}$.

**Theorem 5.7.** Suppose that $F(\mathbf{x}) = \mathbb{E}_{\xi \in \mathcal{D}} F(\mathbf{x}; \xi)$ has $\sigma^2$-sub-Gaussian stochastic gradient, where each $F(\mathbf{x}; \xi)$ is $L_1$-smooth and $L_2$-Hessian Lipschitz continuous. Let $0 < \epsilon, \epsilon_H < 1$ and

$$B_0 = \sigma^2 \epsilon^{-2} \cdot \max\left\{64\left(1 + \log\left[2500 C_1 \max\left\{54\sigma^2 L_1^{-1} L_2^2 \epsilon_H^{-3}, 6\right\} \Delta_F L_1 \epsilon^{-2}\right]\right), 96 C_1\right\}, \quad (5.3)$$

where $C_1 = 200$. Define $\rho = \max\{54\sigma^2 L_1^{-1} L_2^2 \epsilon_H^{-3} B_0^{-1/2}, 6\}$. Set $\delta = 1/(3000 \Delta_F L_2^2 \epsilon_H^{-3})$, the number of epochs $U = 216 \Delta_F L_2^2 \epsilon_H^{-3} + 96 C_1 \rho \Delta_F L_1 B_0^{-1/2} \epsilon^{-2}$, the step size of Algorithm 3 $\eta = \epsilon_H / L_2$ and the step size of One-epoch-SNVRG$^+$ $M = 2\rho L_1$. Choose all the rest parameters of One-epoch-SNVRG$^+$ as in Lemma 5.1. Then with probability at least $1/4$, SNVRG$^+$ + Neon2$^{\text{online}}$ will find an $(\epsilon, \epsilon_H)$-second-order stationary point within

$$\widetilde{O}\left(\frac{\Delta_F L_1^2 L_2^2}{\epsilon_H^5} + \frac{\Delta_F \sigma^2 L_2^2}{\epsilon_H^3 \epsilon^2} + \frac{\Delta_F \sigma L_1}{\epsilon^3}\right) \quad (5.4)$$

stochastic gradient evaluations.

The gradient complexity in Theorem 5.7 again holds with constant probability $1/4$ and we can boost it to a high probability using the same trick as we discussed in Remark 5.5.

**Remark 5.8.** Theorem 5.7 suggests that the gradient complexity of Algorithm 3 is $\widetilde{O}(\epsilon^{-3} + \epsilon_H^{-5} + \epsilon^{-2}\epsilon_H^{-3})$. In contrast, the gradient complexity of SCSG+Neon2$^{\text{online}}$ (Allen-Zhu and Li, 2017) is $\widetilde{O}(\epsilon^{-10/3} + \epsilon_H^{-5} + \epsilon^{-2}\epsilon_H^{-3})$ and that of Natasha2+Neon2$^{\text{online}}$ (Allen-Zhu, 2017) is $\widetilde{O}(\epsilon^{-3.25} + \epsilon_H^{-5} + \epsilon^{-3}\epsilon_H)$. Our algorithm is evidently faster than these two algorithms in the first term in the big O notation. We visualize the gradient complexities of these three algorithms in Figure 2. To better visualize the differences, we divide the entire regime of $\epsilon_H$ into two regimes: (a) $\epsilon_H < \sqrt{\epsilon}$ and (b) $\epsilon_H \geq \sqrt{\epsilon}$, and plot them separately in Figures 2(a) and 2(b). From Figure 2, we have the following discussion.

- When $\epsilon_H \leq \epsilon$, all three algorithms achieve $\widetilde{O}(\epsilon_H^{-5})$ gradient complexity.

- When $\epsilon < \epsilon_H < \sqrt{\epsilon}$, both SNVRG$^+$ + Neon2$^{\text{online}}$ and SCSG+Neon2$^{\text{online}}$ attain $\widetilde{O}(\epsilon^{-2}\epsilon_H^{-3})$ gradient complexity and are worse than Natasha2+Neon2$^{\text{online}}$, which has $\widetilde{O}(\epsilon_H^{-5})$ gradient complexity for $\epsilon_H \in (\epsilon, \epsilon^{3/4})$ and $\widetilde{O}(\epsilon^{-3}\epsilon_H^{-1})$ gradient complexity for $\epsilon_H \in (\epsilon^{3/4}, \epsilon^{1/2})$.

- When $\sqrt{\epsilon} \leq \epsilon_H \leq \epsilon^{4/9}$, SNVRG$^+$ + Neon2$^{\text{online}}$ and SCSG+Neon2$^{\text{online}}$ still attain $\widetilde{O}(\epsilon^{-2}\epsilon_H^{-3})$ gradient complexity but perform better than Natasha2+Neon2$^{\text{online}}$, which has $\widetilde{O}(\epsilon^{-3}\epsilon_H^{-1})$ gradient complexity.

- When $\epsilon_H \geq \epsilon^{4/9}$, SNVRG$^+$+Neon2$^{\text{online}}$ beats both SCSG+Neon2$^{\text{online}}$ and Natasha2+Neon2$^{\text{online}}$.

In particular, when $\epsilon_H = \epsilon^{1/3}$, our algorithm SNVRG$^+$+Neon2$^{\text{online}}$ is faster than SCSG+Neon2$^{\text{online}}$ and Natasha2+Neon2$^{\text{online}}$ by a factor of $O(\epsilon^{1/3})$. And when $\epsilon_H \geq \epsilon^{1/4}$, SNVRG$^+$ + Neon2$^{\text{online}}$ is faster than Natasha2+Neon2$^{\text{online}}$ by a factor of $O(\epsilon^{1/4})$.

## 6 Main Theory with Third-order Smoothness

As we mentioned before, it has been shown that the third-order smoothness of the objective function $F$ can help accelerate the convergence of nonconvex optimization (Carmon et al., 2017; Yu et al., 2017a). For the intuition of the acceleration by third-order smoothness, we refer readers to the detailed exhibition and discussion in Yu et al. (2017a). In this section, we will show that our local minimum finding algorithms (Algorithms 2 and 3) can find local minima faster provided this additional condition.

### 6.1 Finite-Sum Optimization Problems

We first consider the finite-sum optimization problem in (1.3). The following theorem spells out the gradient complexity of Algorithm 2 under additional third-order smoothness.

**Theorem 6.1.** Suppose that $F = 1/n \sum_{i=1}^n f_i$, where each $f_i$ is $L_1$-smooth, $L_2$-Hessian Lipschitz continuous and $F$ is $L_3$-third-order smooth. Let $0 < \epsilon, \epsilon_H < 1$, $\delta = \epsilon_H^2/(72L_3\Delta_F)$ and $U = 12L_3\Delta_F\epsilon_H^{-2} + 1800CL_1\Delta_F\epsilon^{-2}n^{-1/2}$. Set $B_0 = n, M = 6L_1$ and all the rest parameters of One-epoch-SNVRG$^+$ as in Lemma 5.1. Choose the step size as $\eta = \sqrt{3\epsilon_H/L_3}$. Then with probability at least $1/4$, SNVRG$^+$ + Neon2$^{\text{finite}}$ will find an $(\epsilon, \epsilon_H)$-second-order stationary point within

$$\widetilde{O}\bigg(\frac{\Delta_F n L_3}{\epsilon_H^2} + \frac{\Delta_F n^{3/4} L_1^{1/2} L_3}{\epsilon_H^{5/2}} + \frac{\Delta_F n^{1/2} L_1}{\epsilon^2}\bigg) \quad (6.1)$$

stochastic gradient evaluations.

Similar to previous discussions, we can repeatedly run Algorithm 2 for $\log(1/p)$ times to boost its confidence to $1 - p$ for any $p \in (0, 1)$.

**Remark 6.2.** Compared with step size $\eta = \epsilon_H/L_2$ used in the negative curvature descent step (Line 11) of Algorithm 2 in Theorem 5.4 without third-order smoothness, the step size in Theorem 6.1 is chosen to be $\eta = \sqrt{\epsilon_H/L_3}$ where $L_3$ is the third-order smoothness parameter. Note that when $\epsilon_H \ll 1$, the step size we choose under third-order smoothness assumption is much bigger than that under only second-order smoothness assumption. As is pointed out by Yu et al. (2017a), the key advantage of third-order smoothness condition is that it enables us to choose a larger step size and therefore achieve much more function value decrease in the negative curvature descent step (Line 11 of Algorithm 2).

**Remark 6.3.** Theorem 6.1 suggests that the gradient complexity of $\text{SNVRG}^+ + \text{Neon2}^{\text{finite}}$ under third-order smoothness is $\widetilde{O}(n^{1/2}\epsilon^{-2} + n\epsilon_H^{-2} + n^{3/4}\epsilon_H^{-5/2})$. In stark contrast, the gradient complexity of the state-of-the-art finite-sum local minimum finding algorithm with third-order smoothness assumption (FLASH) (Yu et al., 2017a) is $\widetilde{O}(n^{2/3}\epsilon^{-2} + n\epsilon_H^{-2} + n^{3/4}\epsilon_H^{-5/2})$. Clearly, our algorithm is strictly better than the FLASH algorithm (Yu et al., 2017a) in the first term of the gradient complexity.

Specifically, if we choose $\epsilon_H = \sqrt{\epsilon}$, $\text{SNVRG}^+ + \text{Neon2}^{\text{finite}}$ is faster for finding an $(\epsilon, \sqrt{\epsilon})$-approximate local minimum than FLASH by a factor of $O(1/\epsilon^{1/6})$ when $n \lesssim \epsilon^{-2}$. $\text{SNVRG}^+ + \text{Neon2}^{\text{finite}}$ is also strictly faster than FLASH when $\epsilon^{-2} \lesssim n \lesssim \epsilon^{-3}$ and will match FLASH when $n \gtrsim \epsilon^{-3}$. We show this comparison in Figure 3, which clearly demonstrates that the gradient complexity of $\text{SNVRG}^+ + \text{Neon2}^{\text{finite}}$ is much smaller than that of FLASH in a very wide regime.

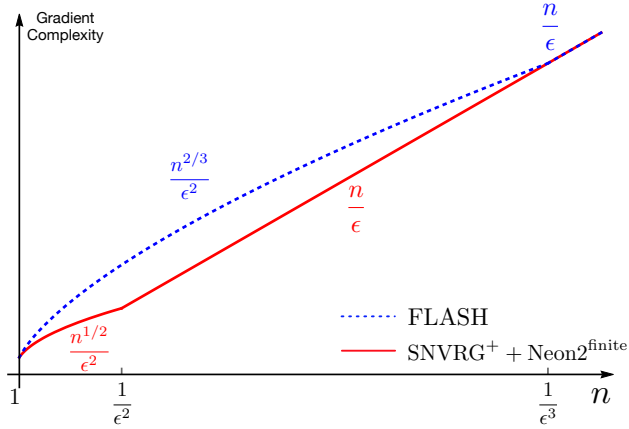

Figure 3: Comparison of gradient complexities between $\text{SNVRG}^+ + \text{Neon2}^{\text{finite}}$ and FLASH for finding an $(\epsilon, \sqrt{\epsilon})$-approximate local minimum in finite-sum nonconvex optimization problems.

## 6.2 General Stochastic Optimization Problems

Now we turn to the general stochastic optimization problem (1.4). We characterize the gradient complexity of Algorithm 3 under third-order smoothness in the following theorem.

**Theorem 6.4.** Suppose that $F(\mathbf{x}) = \mathbb{E}_{\xi \in \mathcal{D}} F(\mathbf{x}; \xi)$ has $\sigma^2$-sub-Gaussian stochastic gradient, where each $F(\mathbf{x}; \xi)$ is $L_1$-smooth, $L_2$-Hessian Lipschitz continuous and $F(\mathbf{x})$ is $L_3$-third-order smooth. Let $0 < \epsilon, \epsilon_H < 1$, and

$$B_0 = \sigma^2 \epsilon^{-2} \cdot \max\left\{64\left(1 + \log\left[2500 C_1 \max\{36\sigma^2 L_1^{-1} L_3 \epsilon_H^{-2}, 6\} \Delta_F L_1 \epsilon^{-2}\right]\right), 96 C_1\right\}, \quad (6.2)$$

where $C_1 = 200$. Define $\rho = \max\{36\sigma^2 L_1^{-1} L_3 \epsilon_H^{-2} B_0^{-1/2}, 6\}$. Let $\delta = 1/(1000 \Delta_F L_3 \epsilon_H^{-2})$, the number of epochs $U = 72 \Delta_F L_3 \epsilon_H^{-2} + 96 C_1 \rho \Delta_F L_1 B_0^{-1/2} \epsilon^{-2}$, the step size $\eta = \sqrt{\epsilon_H/L_3}$ and the step size of One-epoch-SNVRG$^+$ $M = 2\rho L_1$. Choose all the rest parameters of One-epoch-SNVRG$^+$ the same as in Lemma 5.1. Then with probability at least $1/4$, SNVRG$^+$ + Neon2$^{\text{online}}$ will find an $(\epsilon, \epsilon_H)$-second-order stationary point within

$$\widetilde{O}\bigg(\frac{\Delta_F L_1^2 L_3}{\epsilon_H^4} + \frac{\Delta_F \sigma^2 L_3}{\epsilon_H^2 \epsilon^2} + \frac{\Delta_F \sigma L_1}{\epsilon^3}\bigg) \qquad (6.3)$$

stochastic gradient evaluations.

**Remark 6.5.** Theorem 6.4 suggests that the gradient complexity of Algorithm 3 under third-order smoothness is $\widetilde{O}(\epsilon^{-3} + \epsilon_H^{-4} + \epsilon^{-2}\epsilon_H^{-2})$. As a comparison, the gradient complexity of existing best stochastic local minimum finding algorithm with third-order smoothness assumption (FLASH) (Yu et al., 2017a) is $\widetilde{O}(\epsilon^{-10/3} + \epsilon_H^{-4} + \epsilon^{-2}\epsilon_H^{-2})$. The gradient complexity of SNVRG$^+$ + Neon2$^{\text{online}}$ is faster than that of FLASH in the first term. We illustrate the comparison of gradient complexities of both algorithms in Figure 4. It is evident that when $\epsilon_H \geq \epsilon^{2/3}$, our algorithm SNVRG$^+$+Neon2$^{\text{online}}$ always enjoys a lower gradient complexity than FLASH. In addition, if we choose $\epsilon_H = \sqrt{\epsilon}$, our algorithm is faster for finding an $(\epsilon, \sqrt{\epsilon})$-approximate local minimum than FLASH (Yu et al., 2017a) by a factor of $O(\epsilon^{-1/3})$.

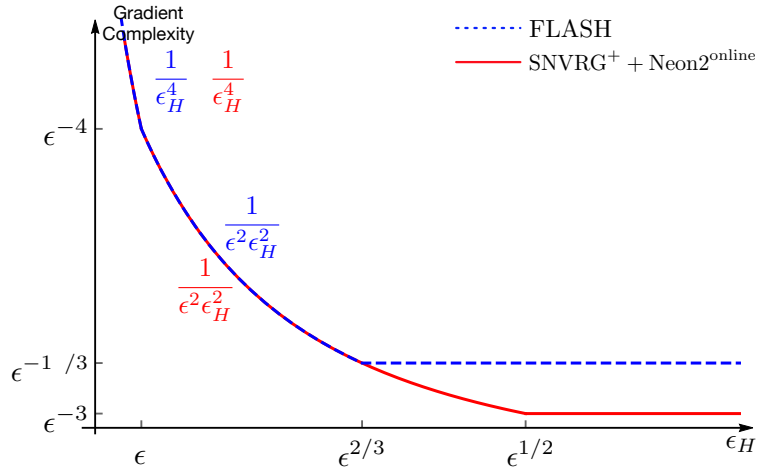

Figure 4: Comparison of gradient complexities between SNVRG$^+$ + Neon2$^{\text{online}}$ and FLASH for finding an $(\epsilon, \epsilon_H)$-approximate second-order stationary point in general stochastic problems.

## 7 Conclusions

In this paper, we proposed two algorithms that can find local minima of nonconvex optimization problems faster than existing algorithms in the literature. The proposed algorithms are based on One-epoch-SNVRG$^+$ which uses stochastic nested variance reduction techniques and outperform state-of-the-art algorithms in terms of gradient complexity in both finite-sum optimization and general stochastic optimization. We further extended the analyses of our algorithms to the case where the objective function is third-order smooth and proved a even faster convergence to a local minimum under this additional assumption.

# A Proof of the Main Theory

In this section, we provide the proofs of gradient complexities of our proposed algorithms SNVRG$^+$ + Neon2$^{\text{finite}}$ and SNVRG$^+$ + Neon2$^{\text{online}}$. To avoid distracting readers from the proofs of our main results, we defer the proof of Lemma 5.1 to Appendix C.

## A.1 Proof of Theorem 5.4

When $F(\mathbf{x})$ has the finite-sum structure in (1.3), we choose $B_0 = n$, $M = 6L_1$ in One-epoch-SNVRG$^+$. Based on Lemma 5.1 we have the following corollary.

**Corollary A.1.** Suppose that each $f_i$ is $L_1$-smooth. We choose $B_0 = n$, and let other parameters be chose as in Lemma 5.1. Then the output of Algorithm 1 $\mathbf{x}_T$ satisfies

$$\mathbb{E}\|\nabla F(\mathbf{x}_T)\|_2^2 \leq \frac{CL_1}{n^{1/2}} \cdot \mathbb{E}\big[F(\mathbf{x}_0) - F(\mathbf{x}_T)\big], \tag{A.1}$$

where $C = 600$. Let $\mathcal{T}$ be the total amount of stochastic gradient computations of Algorithm 1, then we have $\mathbb{E}\mathcal{T} \leq 20n \log^3 n$.

The following lemma shows that based on Neon2$^{\text{finite}}$ the negative curvature descent step of Algorithm 2 (Line 11) enjoys sufficient function value decrease. The proof can be found in Theorem 5 and Claim C.2 in Allen-Zhu and Li (2017).

**Lemma A.2.** (Allen-Zhu and Li, 2017) Suppose that $F = 1/n \sum_{i=1}^{n} f_i$, each $f_i$ is $L_1$-smooth and $L_2$-Hessian Lipschitz continuous. Let $\epsilon_H \in (0,1)$ and set $\eta = \epsilon_H/L_2$. Suppose that $\lambda_{\min}(\nabla^2 F(\mathbf{z}_{u-1})) < -\epsilon_H$ and that at the $u$-th iteration Algorithm 2 executes the Neon2$^{\text{finite}}$ algorithm (Line 6). Then with probability $1-\delta$ it holds that

$$\mathbb{E}_\zeta\big[F(\mathbf{z}_u) - F(\mathbf{z}_{u-1})\big] \leq -\epsilon_H^3/(12L_2^2). \tag{A.2}$$

In addition, Neon2$^{\text{finite}}$ takes $O\big((n + n^{3/4}\sqrt{L_1/\epsilon_H})\log^2(d/\delta)\big)$ stochastic gradient computations.

*Proof of Theorem 5.4.* Let $\mathcal{I} = \{1, \ldots, U\}$ be the index set of all iterations. We denote $\mathcal{I}_1$ and $\mathcal{I}_2$ as the index sets such that $\mathbf{z}_u$ is obtained from Neon2$^{\text{finite}}$ for all $u \in \mathcal{I}_1$ and $\mathbf{z}_{u'}$ is the output by SNVRG$^+$ for all $u' \in \mathcal{I}_2$. Obviously we have $U = |\mathcal{I}_1| + |\mathcal{I}_2|$. We will calculate $|\mathcal{I}_1|, |\mathcal{I}_2|$ separately. For $|\mathcal{I}_1|$, by Lemma A.2, with probability $1-\delta$, we have

$$\mathbb{E}\big[F(\mathbf{z}_u) - F(\mathbf{z}_{u-1})\big] \leq -\epsilon_H^3/(12L_2^2), \qquad \text{for } u \in \mathcal{I}_1. \tag{A.3}$$

Summing up (A.3) over $u \in \mathcal{I}_1$, then with probability $1 - \delta \cdot |\mathcal{I}_1|$ we have

$$|\mathcal{I}_1| \cdot \epsilon_H^3/(12L_2^2) \leq \sum_{u \in \mathcal{I}_1} \mathbb{E}\big[F(\mathbf{z}_{u-1}) - F(\mathbf{z}_u)\big] \leq \sum_{u \in \mathcal{I}} \mathbb{E}\big[F(\mathbf{z}_{u-1}) - F(\mathbf{z}_u)\big] \leq \Delta_F, \tag{A.4}$$

where the second inequality holds because by Corollary A.1 it holds that

$$0 \leq \mathbb{E}\|\nabla F(\mathbf{z}_u)\|_2^2 \leq \frac{CL_1}{n^{1/2}}\mathbb{E}\big[F(\mathbf{z}_{u-1}) - F(\mathbf{z}_u)\big], \qquad \text{for all } u \in \mathcal{I}_2. \tag{A.5}$$

By (A.4), we have
$$|\mathcal{I}_1| \leq 12L_2^2 \Delta_F / \epsilon_H^3. \tag{A.6}$$

To calculate $|\mathcal{I}_2|$, we further decompose $\mathcal{I}_2$ into two disjoint sets such that $\mathcal{I}_2 = \mathcal{I}_2^1 \cup \mathcal{I}_2^2$, where $\mathcal{I}_2^1 = \{u \in \mathcal{I}_2 : \|\mathbf{g}_u\|_2 > \epsilon\}$, $\mathcal{I}_2^2 = \{u \in \mathcal{I}_2 : \|\mathbf{g}_u\|_2 \leq \epsilon\}$. It is worth noting that if $u \in \mathcal{I}_2^2$ such that $\|\mathbf{g}_u\|_2 \leq \epsilon$, then Algorithm 2 will execute Neon2$^{\text{finite}}$ and a negative curvature descent step, which means $u+1 \in \mathcal{I}_1$ by definition. Thus, it always holds that $|\mathcal{I}_2^2| \leq |\mathcal{I}_1|$. For $|\mathcal{I}_2^1|$, note that $\mathbf{x}_0 = \mathbf{z}_{u-1}$ and $\mathbf{x}_T = \mathbf{z}_u$ in Corollary A.1, which directly implies
$$\sum_{u \in \mathcal{I}_2^1} \mathbb{E}\|\nabla F(\mathbf{z}_u)\|_2^2 \leq \sum_{u \in \mathcal{I}_2^1} \frac{CL_1}{n^{1/2}} \mathbb{E}[F(\mathbf{z}_{u-1}) - F(\mathbf{z}_u)]$$
$$\leq \sum_{u \in \mathcal{I}} \frac{CL_1}{n^{1/2}} \mathbb{E}[F(\mathbf{z}_{u-1}) - F(\mathbf{z}_u)]$$
$$\leq \frac{CL_1}{n^{1/2}} \cdot \Delta_F, \tag{A.7}$$

where the second inequality holds because $\mathbb{E}[F(\mathbf{z}_{u-1}) - F(\mathbf{z}_u)] \geq 0$ for $u \in \mathcal{I}_1 \cup \mathcal{I}_2$ by (A.3) and (A.5). Applying Markov's inequality, with probability at least $2/3$, we have
$$\sum_{u \in \mathcal{I}_2^1} \|\nabla F(\mathbf{z}_u)\|_2^2 \leq \frac{3CL_1\Delta_F}{n^{1/2}}. \tag{A.8}$$

Since for any $u \in \mathcal{I}_2^1$, we have $\|\nabla F(\mathbf{z}_u)\|_2 = \|\mathbf{g}_u\|_2 > \epsilon$, with probability at least $2/3$ it holds that
$$|\mathcal{I}_2^1| \leq \frac{3CL_1\Delta_F}{\epsilon^2 n^{1/2}}. \tag{A.9}$$

Thus, the total number of iterations is $U = |\mathcal{I}_1| + |\mathcal{I}_2| \leq 2|\mathcal{I}_1| + |\mathcal{I}_2^1| \leq 24L_2^2\Delta_F\epsilon_H^{-3} + 3CL_1\Delta_F\epsilon^{-2}n^{-1/2}$.

We now calculate the gradient complexity of Algorithm 2. By Corollary A.1 one single call of One-epoch-SNVRG$^+$ needs at most $20n\log^3 n$ stochastic gradient computations and by Lemma A.2 one single call of Neon2$^{\text{finite}}$ needs $O\big((n + n^{3/4}\sqrt{L_1/\epsilon_H})\log^2(d/\delta)\big)$ stochastic gradient computations. In addition, we need to compute $\mathbf{g}_u$ at each iteration of Algorithm 2 (Line 2), which takes $O(n)$ stochastic gradient computations. Thus, the expectation of the total amount of stochastic gradient computations, denoted by $\mathbb{E}T_{\text{total}}$, can be upper bounded by
$$|\mathcal{I}_1| \cdot O\big((n + n^{3/4}\sqrt{L_1/\epsilon_H})\log^2(d/\delta)\big) + |\mathcal{I}_2| \cdot O(n\log^3 n) + |\mathcal{I}| \cdot O(n)$$
$$= |\mathcal{I}_1| \cdot \widetilde{O}\big(n + n^{3/4}\sqrt{L_1/\epsilon_H}\big) + (|\mathcal{I}_2^1| + |\mathcal{I}_2^2|) \cdot \widetilde{O}(n)$$
$$= |\mathcal{I}_1| \cdot \widetilde{O}\big(n + n^{3/4}\sqrt{L_1/\epsilon_H}\big) + (|\mathcal{I}_2^1| + |\mathcal{I}_1|) \cdot \widetilde{O}(n). \tag{A.10}$$

We further plug the upper bound for $|\mathcal{I}_1|$ and $|\mathcal{I}_2^1|$ into (A.10) and obtain
$$\mathbb{E}T_{\text{total}} = O(L_2^2\Delta_F\epsilon_H^{-3}) \cdot \widetilde{O}\big(n + n^{3/4}\sqrt{L_1/\epsilon_H}\big) + O(L_1\Delta_F\epsilon^{-2}n^{-1/2})\widetilde{O}(n)$$
$$= \widetilde{O}\big(\Delta_F n L_2^2 \epsilon_H^{-3} + \Delta_F n^{3/4} L_1^{1/2} L_2^2 \epsilon_H^{-7/2} + \Delta_F n^{1/2} L_1 \epsilon^{-2}\big). \tag{A.11}$$

Finally, applying Markov inequality, with probability $2/3$, it holds that
$$T_{\text{total}} = \widetilde{O}\big(\Delta_F n L_2^2 \epsilon_H^{-3} + \Delta_F n^{3/4} L_1^{1/2} L_2^2 \epsilon_H^{-7/2} + \Delta_F n^{1/2} L_1 \epsilon^{-2}\big). \tag{A.12}$$

Since $|\mathcal{I}_1|\delta = |\mathcal{I}_1|/(144 \cdot L_2^2 \Delta_F \epsilon_H^{-3}) \leq 1/12$, then by the union bound, with probability $1 - 1/3 - 1/3 - |\mathcal{I}_1|\delta \geq 1/4$, SNVRG$^+$ + Neon2$^{\text{finite}}$ will find an $(\epsilon, \epsilon_H)$-second order stationary point within

$$\widetilde{O}\big(\Delta_F n L_2^2 \epsilon_H^{-3} + \Delta_F n^{3/4} L_1^{1/2} L_2^2 \epsilon_H^{-7/2} + \Delta_F n^{1/2} L_1 \epsilon^{-2}\big) \tag{A.13}$$

stochastic gradient computations. □

## A.2 Proof of Theorem 5.7

For general stochastic problem in (1.4), we denote $\nabla F(\mathbf{x}; \xi_i)$ as one subsampled gradient at $\mathbf{x}$ for any random variable $\xi$ and index $i$ in One-epoch-SNVRG$^+$. Base on Lemma 5.1 we have the next corollary.

**Corollary A.3.** Suppose that for each $\xi$, $F(\mathbf{x}; \xi)$ is $L_1$-smooth and has $\sigma^2$-sub-Gaussian stochastic gradient. We choose $M = 2\rho L_1$ and suppose that $n \gg O(\epsilon^{-2})$ and $B_0 < n$. Then the output of Algorithm 1 $\mathbf{x}_T$ satisfies

$$\mathbb{E}\|\nabla F(\mathbf{x}_T)\|_2^2 \leq C_1 \bigg( \frac{\rho L_1}{B_0^{1/2}} \cdot \mathbb{E}\big[F(\mathbf{x}_0) - F(\mathbf{x}_T)\big] + \frac{\sigma^2}{B_0} \bigg), \tag{A.14}$$

where $C_1 = 200$. The total amount of stochastic gradient computations of Algorithm 1 is $\mathbb{E}\mathcal{T} \leq 20 B_0 \log^3 B_0$.

The following lemma shows that based on Neon2$^{\text{online}}$ the negative curvature descent step of Algorithm 3 (Line 12) enjoys sufficient function value decrease. The proof can be found in Lemma 3.1 and Claim C.2 in Allen-Zhu and Li (2017).

**Lemma A.4.** (Allen-Zhu and Li, 2017) Suppose that $F(\mathbf{x}) = \mathbb{E}_{\xi \in \mathcal{D}} F(\mathbf{x}; \xi)$ and each $F(\mathbf{x}; \xi)$ is $L_1$-smooth, $L_2$-Hessian Lipschitz continuous. Let $\epsilon_H \in (0, 1)$ and set $\eta = \epsilon_H/L_2$. Suppose that $\lambda_{\min}(\nabla^2 F(\mathbf{z}_{u-1})) < -\epsilon_H$ and that at the $u$-th iteration Algorithm 3 executes the Neon2$^{\text{online}}$ algorithm (Line 7). Then with probability $1 - \delta$ it holds that

$$\mathbb{E}_{\zeta}\big[F(\mathbf{z}_u) - F(\mathbf{z}_{u-1})\big] \leq -\epsilon_H^3/(12L_2^2). \tag{A.15}$$

In addition, Neon2$^{\text{online}}$ takes $O(L_1^2/\epsilon_H^2 \log^2(d/\delta))$ stochastic gradient computations.

We also need the following concentration inequality in our proof.

**Lemma A.5.** (Ghadimi et al., 2016) Suppose the stochastic gradient $\nabla F(\mathbf{x}; \xi)$ is $\sigma^2$-sub-Gaussian. Let $\nabla F_{\mathcal{S}}(\mathbf{x}) = 1/|\mathcal{S}| \sum_{i \in \mathcal{S}} \nabla F(\mathbf{x}; \xi_i)$, where $\mathcal{S}$ is a subsampled gradient of $F(\mathbf{x})$. If the sample size satisfies $|\mathcal{S}| = 2\sigma^2/\epsilon^2 (1 + \log^{1/2}(1/\delta))^2$, then with probability at least $1 - \delta$,

$$\|\nabla F_{\mathcal{S}}(\mathbf{x}) - \nabla F(\mathbf{x})\|_2 \leq \epsilon.$$

*Proof of Theorem 5.7.* Denote $\delta_0 = 1/(2500 C_1 \rho \Delta_F L_1 B_0^{-1/2} \epsilon^{-2})$. Then by the choice of $B_0$ in (5.3) it holds that $B_0 > 32\sigma^2/\epsilon^2 (1 + \log^{1/2}(1/\delta_0))^2$. Let $\mathcal{I} = \{1, \ldots, U\}$ be the index set of all iterations. We use $\mathcal{I}_1$ and $\mathcal{I}_2$ to represent the index set of iterates where the $\mathbf{z}_u$ is obtained from Neon2$^{\text{finite}}$ and SNVRG$^+$ respectively. From Lemma A.4, we have that with probability at least $1 - \delta$ that

$$\mathbb{E}\big[F(\mathbf{z}_u) - F(\mathbf{z}_{u-1})\big] \leq -\epsilon_H^3/(12L_2^2), \qquad \text{for } u \in \mathcal{I}_1. \tag{A.16}$$

By Corollary A.3, we have

$$\mathbb{E}\|\nabla F(\mathbf{z}_u)\|_2^2 \leq C_1 \left( \frac{\rho L_1}{B_0^{1/2}} \cdot \mathbb{E}\left[F(\mathbf{z}_{u-1}) - F(\mathbf{z}_u)\right] + \frac{\sigma^2}{B_0} \right), \qquad \text{for } u \in \mathcal{I}_2. \tag{A.17}$$

where $C_1 = 200$. We further decompose $\mathcal{I}_2 = \mathcal{I}_2^1 \cup \mathcal{I}_2^2$, where $\mathcal{I}_2^1 = \{u \in \mathcal{I}_2 : \|\mathbf{g}_u\|_2 > \epsilon/2\}$ and $\mathcal{I}_2^1 = \{u \in \mathcal{I}_2 : \|\mathbf{g}_u\|_2 \leq \epsilon/2\}$. (A.17) immediately implies the following two inequalities:

$$\mathbb{E}\left[F(\mathbf{z}_u) - F(\mathbf{z}_{u-1})\right] \leq -\frac{B_0^{1/2}}{C_1 \rho L_1} \mathbb{E}\|\nabla F(\mathbf{z}_u)\|_2^2 + \frac{\sigma^2}{\rho L_1 B_0^{1/2}}, \qquad u \in \mathcal{I}_2^1, \tag{A.18}$$

$$\mathbb{E}\left[F(\mathbf{z}_u) - F(\mathbf{z}_{u-1})\right] \leq \frac{\sigma^2}{\rho L_1 B_0^{1/2}}, \qquad u \in \mathcal{I}_2^2. \tag{A.19}$$

Summing up (A.16) over $u \in \mathcal{I}_1$, (A.18) over $u \in \mathcal{I}_2^1$ and (A.19) over $u \in \mathcal{I}_2^2$, we have

$$\mathbb{E}\left[\sum_{u \in \mathcal{I}} F(\mathbf{z}_{u-1}) - F(\mathbf{z}_u)\right] \geq \frac{|\mathcal{I}_1| \epsilon_H^3}{12 L_2^2} + \frac{B_0^{1/2}}{C_1 \rho L_1} \sum_{u \in \mathcal{I}_2^1} \mathbb{E}\|\nabla F(\mathbf{z}_u)\|_2^2 - \sum_{u \in \mathcal{I}_2^1} \frac{\sigma^2}{\rho L_1 B_0^{1/2}} - \sum_{u \in \mathcal{I}_2^2} \frac{\sigma^2}{\rho L_1 B_0^{1/2}}. \tag{A.20}$$

Since for any $u \in \mathcal{I}_2^2$ we have $\|\mathbf{g}_u\|_2 \leq \epsilon/2$, Algorithm 3 will execute Neon2$^{\text{online}}$ at the $u$-th iteration, which indicates $|\mathcal{I}_2^2| \leq |\mathcal{I}_1|$. Combining this with (A.20) and by the definition of $\Delta_F$, with probability at least $1 - |\mathcal{I}_1|\delta$, we have

$$\frac{|\mathcal{I}_1| \epsilon_H^3}{12 L_2^2} + \frac{B_0^{1/2}}{C_1 \rho L_1} \sum_{u \in \mathcal{I}_2^1} \mathbb{E}\|\nabla F(\mathbf{z}_u)\|_2^2 \leq \Delta_F + (|\mathcal{I}_1| + |\mathcal{I}_2^1|) \frac{\sigma^2}{\rho L_1 B_0^{1/2}}. \tag{A.21}$$

Using Markov inequality, then with probability at least $2/3$, it holds that

$$\frac{|\mathcal{I}_1| \epsilon_H^3}{12 L_2^2} + \frac{B_0^{1/2}}{C_1 \rho L_1} \sum_{u \in \mathcal{I}_2^1} \|\nabla F(\mathbf{z}_u)\|_2^2 \leq 3 \left( \Delta_F + (|\mathcal{I}_1| + |\mathcal{I}_2^1|) \frac{\sigma^2}{\rho L_1 B_0^{1/2}} \right). \tag{A.22}$$

By Lemma A.5, for any $u \in \mathcal{I}_2^1$, with probability at least $1 - \delta_0$, it holds that $\|\nabla F(\mathbf{z}_u) - \mathbf{g}_u\|_2 < \epsilon/4$ if $B_0 \geq 32\sigma^2/\epsilon^2 (1 + \log^{1/2}(1/\delta_0))^2$, which further indicates that $\|\nabla F(\mathbf{z}_u)\|_2 > \epsilon/4$. Thus, applying union bound yields that with probability at least $1 - |\mathcal{I}_1|\delta - 1/3 - |\mathcal{I}_2^1|\delta_0$ we have

$$\frac{|\mathcal{I}_1| \epsilon_H^3}{12 L_2^2} + \frac{|\mathcal{I}_2^1| B_0^{1/2} \epsilon^2}{16 C_1 \rho L_1} \leq 3\Delta_F + \frac{3|\mathcal{I}_2^1| \sigma^2}{\rho L_1 B_0^{1/2}} + \frac{3|\mathcal{I}_1| \sigma^2}{\rho L_1 B_0^{1/2}}. \tag{A.23}$$

Recall that in Theorem 5.7 we set $\rho = \max\{54\sigma^2 L_1^{-1} L_2^2 \epsilon_H^{-3} B_0^{-1/2}, 6\} \geq 54\sigma^2 L_2^2/(L_1 \epsilon_H^3 B_0^{1/2})$, which implies

$$\frac{3|\mathcal{I}_1| \sigma^2}{\rho L_1 B_0^{1/2}} \leq \frac{|\mathcal{I}_1| \epsilon_H^3}{18 L_2^2}. \tag{A.24}$$

By (5.3) we have $B_0 > 96C_1\sigma^2\epsilon^{-2}$, which implies

$$\frac{3|\mathcal{I}_2^1|\sigma^2}{\rho L_1 B_0^{1/2}} \leq \frac{|\mathcal{I}_2^1|B_0^{1/2}\epsilon^2}{32C_1\rho L_1}. \tag{A.25}$$

Plugging (A.24) and (A.25) into (A.23) and rearranging the resulting inequality, then with probability $1 - |\mathcal{I}_1|\delta - 1/3 - |\mathcal{I}_2^1|\delta_0$, we have

$$\frac{|\mathcal{I}_1|\epsilon_H^3}{36L_2^2} + \frac{|\mathcal{I}_2^1|B_0^{1/2}\epsilon^2}{32C_1\rho L_1} \leq 3\Delta_F,$$

which immediately implies that

$$|\mathcal{I}_1| \leq 108\Delta_F L_2^2 \epsilon_H^{-3} = O(\Delta_F L_2^2 \epsilon_H^{-3}), \tag{A.26}$$

and

$$\begin{aligned}|\mathcal{I}_2^1| &\leq 96C_1\rho\Delta_F L_1 B_0^{-1/2}\epsilon^{-2}\\ &= \max\{54\sigma^2 L_1^{-1} L_2^2 \epsilon_H^{-3} B_0^{-1/2}, 6\} \cdot 96 C_1 \Delta_F L_1 B_0^{-1/2}\epsilon^{-2}\\ &= \widetilde{O}(\Delta_F \sigma^{-1} L_1 \epsilon^{-1}) + \widetilde{O}(\Delta_F L_2^2 \epsilon_H^{-3}).\end{aligned} \tag{A.27}$$

Thus we can calculate $U = |\mathcal{I}_1| + |\mathcal{I}_2| \leq 2|\mathcal{I}_1| + |\mathcal{I}_2^1| \leq 216\Delta_F L_2^2 \epsilon_H^{-3} + 96C_1\rho\Delta_F L_1 B_0^{-1/2}\epsilon^{-2}$. Now we are ready to calculate the gradient complexity of Algorithm 3. By Lemma A.4, we know that one single call of Neon2$^{\text{online}}$ needs $O(L_1^2/\epsilon_H^2 \log^2(d/\delta))$ stochastic gradient computations, and one single call of One-epoch-SNVRG$^+$ needs $20B_0 \log^3 B_0 = \widetilde{O}(\sigma^2/\epsilon^2)$ stochastic gradient computations. In addition, we need to compute $\mathbf{g}_u$ at each iteration of Algorithm 3 (Line 2), which costs $B_0 = \widetilde{O}(\sigma^2/\epsilon^2)$ stochastic gradient computations. Thus, the expected total amount of stochastic gradient computations $\mathbb{E}T_{\text{total}}$ can be bounded as

$$\begin{aligned}\mathbb{E}T_{\text{total}} &= |\mathcal{I}_1| \cdot O(L_1^2/\epsilon_H^2 \log^2(d/\delta)) + |\mathcal{I}_2| \cdot \widetilde{O}(\sigma^2/\epsilon^2) + |\mathcal{I}| \cdot \widetilde{O}(\sigma^2/\epsilon^2)\\ &= |\mathcal{I}_1| \cdot \widetilde{O}(L_1^2/\epsilon_H^2) + (|\mathcal{I}_2^1| + |\mathcal{I}_2^2|) \cdot \widetilde{O}(\sigma^2/\epsilon^2) + |\mathcal{I}| \cdot \widetilde{O}(\sigma^2/\epsilon^2)\\ &= |\mathcal{I}_1| \cdot \widetilde{O}(L_1^2/\epsilon_H^2) + (|\mathcal{I}_2^1| + |\mathcal{I}_1|) \cdot \widetilde{O}(\sigma^2/\epsilon^2)\\ &= \widetilde{O}(\Delta_F L_1^2 L_2^2 \epsilon_H^{-5} + \Delta_F \sigma^2 L_2^2 \epsilon_H^{-3}\epsilon^{-2} + \Delta_F \sigma L_1 \epsilon^{-3}).\end{aligned} \tag{A.28}$$

Applying Markov inequality yields

$$T_{\text{total}} = \widetilde{O}(\Delta_F L_1^2 L_2^2 \epsilon_H^{-5} + \Delta_F \sigma^2 L_2^2 \epsilon_H^{-3}\epsilon^{-2} + \Delta_F \sigma L_1 \epsilon^{-3})$$

with probability at least 2/3. Furthermore, we have $|\mathcal{I}_1|\delta = |\mathcal{I}_1|/(3000\Delta_F L_2^2 \epsilon_H^{-3}) \leq 1/24$ and $|\mathcal{I}_2^1|\delta_0 = |\mathcal{I}_2^1|/(2500C_1\rho\Delta_F L_1 B_0^{-1/2}\epsilon^{-2}) < 1/24$. Therefore, with probability at least $1 - |\mathcal{I}_1|\delta - 1/3 - |\mathcal{I}_2^1|\delta_0 - 1/3 \geq 1/4$, Algorithm 3 can find an $(\epsilon, \epsilon_H)$-second order stationary point within

$$\widetilde{O}(\Delta_F L_1^2 L_2^2 \epsilon_H^{-5} + \Delta_F \sigma^2 L_2^2 \epsilon_H^{-3}\epsilon^{-2} + \Delta_F \sigma L_1 \epsilon^{-3})$$

stochastic gradient computations. $\square$

# B Proof of the Main Theory with Third-order Smoothness

In this section, we prove the theoretical results of our proposed algorithms under third-order smoothness condition.

## B.1 Proof of Theorem 6.1

The following lemma shows that the negative curvature descent step (Line 11) of Algorithm 2 achieves more function value decrease under third-order smoothness assumption. The proof can be found in Lemma 4.3 of Yu et al. (2017a).

**Lemma B.1.** (Yu et al., 2017a) Suppose that $F = 1/n \sum_{i=1}^n f_i$, each $f_i$ is $L_1$-smooth, $L_2$-Hessian Lipschitz continuous and $F$ is $L_3$-third-order smooth. Let $\epsilon_H \in (0,1)$ and $\eta = \sqrt{3\epsilon_H/L_3}$. Suppose that $\lambda_{\min}(\nabla^2 F(\mathbf{z}_{u-1})) < -\epsilon_H$ and that at the $u$-th iteration Algorithm 2 executes the Neon2$^{\text{finite}}$ algorithm (Line 6). Then with probability $1 - \delta$ it holds that

$$\mathbb{E}_\zeta[F(\mathbf{z}_u) - F(\mathbf{z}_{u-1})] \leq -\epsilon_H^2/(6L_3). \tag{B.1}$$

In addition, Neon2$^{\text{finite}}$ takes $O\big((n + n^{3/4}\sqrt{L_1/\epsilon_H})\log^2(d/\delta)\big)$ stochastic gradient computations.

*Proof of Theorem 6.1.* Denote $\mathcal{I} = \{1, \ldots, U\}$ as the index of iteration. Let $\mathcal{I} = \{1, \ldots, U\}$ be the index set of iteration. We use $\mathcal{I}_1$ and $\mathcal{I}_2$ to represent the index set of iterates where the $\mathbf{z}_u$ is obtained from Neon2$^{\text{finite}}$ and One-epoch-SNVRG$^+$. Since $U = |\mathcal{I}_1| + |\mathcal{I}_2|$, we calculate $|\mathcal{I}_1|, |\mathcal{I}_2|$ separately. For $|\mathcal{I}_1|$, by Lemma B.1, with probability at least $1 - \delta$, we have

$$\mathbb{E}[F(\mathbf{z}_u) - F(\mathbf{z}_{u-1})] \leq -\epsilon_H^2/(6L_3), \qquad \text{for } u \in \mathcal{I}_1. \tag{B.2}$$

Summing up (B.2) over $u \in \mathcal{I}_1$ and applying union bound, then with probability at least $1 - \delta \cdot |\mathcal{I}_1|$ we have

$$|\mathcal{I}_1| \cdot \epsilon_H^2/(6L_3) \leq \sum_{u \in \mathcal{I}_1} \mathbb{E}[F(\mathbf{z}_{u-1}) - F(\mathbf{z}_u)] \leq \sum_{u \in \mathcal{I}} \mathbb{E}[F(\mathbf{z}_{u-1}) - F(\mathbf{z}_u)] \leq \Delta_F, \tag{B.3}$$

where the second inequality holds due to the fact that by Corollary A.1 we have

$$0 \leq \mathbb{E}\|\nabla F(\mathbf{z}_u)\|_2^2 \leq \frac{CL_1}{n^{1/2}}\mathbb{E}[F(\mathbf{z}_{u-1}) - F(\mathbf{z}_u)], \qquad \text{for } u \in \mathcal{I}_2. \tag{B.4}$$

(B.3) directly implies

$$|\mathcal{I}_1| \leq 6L_3\Delta_F/\epsilon_H^2. \tag{B.5}$$

For $|\mathcal{I}_2|$, we decompose $\mathcal{I}_2 = \mathcal{I}_2^1 \cup \mathcal{I}_2^2$, where $\mathcal{I}_2^1 = \{u \in \mathcal{I}_2 : \|\mathbf{g}_u\|_2 > \epsilon\}$ and $\mathcal{I}_2^2 = \{u \in \mathcal{I}_2 : \|\mathbf{g}_u\|_2 \leq \epsilon\}$. If $u \in \mathcal{I}_2^2$, then at the $(u+1)$-th iteration, Algorithm 2 will execute Neon2$^{\text{finite}}$. Thus, we have $|\mathcal{I}_2^2| \leq |\mathcal{I}_1|$. For $|\mathcal{I}_2^1|$, note that $\mathbf{x}_0 = \mathbf{z}_{u-1}$ and $\mathbf{x}_T = \mathbf{z}_u$ in Corollary A.1 and summing up over $u \in \mathcal{I}_2^1$ yields

$$\sum_{u \in \mathcal{I}_2^1} \mathbb{E}\|\nabla F(\mathbf{z}_u)\|_2^2 \leq \sum_{u \in \mathcal{I}_2^1} \frac{CL_1}{n^{1/2}}\mathbb{E}[F(\mathbf{z}_{u-1}) - F(\mathbf{z}_u)]$$

$$\leq \sum_{u \in \mathcal{I}} \frac{CL_1}{n^{1/2}} \mathbb{E}\big[F(\mathbf{z}_{u-1}) - F(\mathbf{z}_u)\big]$$
$$\leq \frac{CL_1}{n^{1/2}} \cdot \Delta_F, \tag{B.6}$$

where the second inequality follows from (B.3) and (B.4). Applying Markov's inequality, with probability at least 2/3, we have

$$\sum_{u \in \mathcal{I}_2^1} \|\nabla F(\mathbf{z}_u)\|_2^2 \leq \frac{3CL_1 \Delta_F}{n^{1/2}}. \tag{B.7}$$

by definition for any $u \in \mathcal{I}_2^1$, we have $\|\nabla F(\mathbf{z}_u)\|_2 = \|\mathbf{g}_u\|_2 > \epsilon$. Then we have with probability at least 2/3 that

$$|\mathcal{I}_2^1| \leq \frac{3CL_1 \Delta_F}{\epsilon^2 n^{1/2}}. \tag{B.8}$$

Total number of iteration is $U = |\mathcal{I}_1| + |\mathcal{I}_2| \leq 2|\mathcal{I}_1| + |\mathcal{I}_2^1| \leq 12L_3 \Delta_F \epsilon_H^{-2} + 3CL_1 \Delta_F \epsilon^{-2} n^{-1/2}$. We now calculate the gradient complexity of Algorithm 2. By Lemma B.1 one single call of Neon2$^{\text{finite}}$ needs $O\big((n + n^{3/4}\sqrt{L_1/\epsilon_H})\log^2(d/\delta)\big)$ stochastic gradient computations and by Corollary A.1 one single call of One-epoch-SNVRG$^+$ needs $20n \log^3 n$ stochastic gradient computations. Moreover, we need to compute $\mathbf{g}_u$ at each iteration, which takes $O(n)$ stochastic gradient computations. Thus, the expectation of the total amount of stochastic gradient computations $\mathbb{E}T_{\text{total}}$ can be bounded by

$$|\mathcal{I}_1| \cdot O\big((n + n^{3/4}\sqrt{L_1/\epsilon_H})\log^2(d/\delta)\big) + |\mathcal{I}_2| \cdot O(n \log^3 n) + |\mathcal{I}| \cdot O(n)$$
$$= |\mathcal{I}_1| \cdot \widetilde{O}\big(n + n^{3/4}\sqrt{L_1/\epsilon_H}\big) + (|\mathcal{I}_2^1| + |\mathcal{I}_2^2|) \cdot \widetilde{O}(n)$$
$$= |\mathcal{I}_1| \cdot \widetilde{O}\big(n + n^{3/4}\sqrt{L_1/\epsilon_H}\big) + (|\mathcal{I}_2^1| + |\mathcal{I}_1|) \cdot \widetilde{O}(n). \tag{B.9}$$

We further plug the upper bound of $|\mathcal{I}_1|$ and $|\mathcal{I}_2^1|$ into (B.9) and obtain

$$\mathbb{E}T_{\text{total}} = O(L_3 \Delta_F \epsilon_H^{-2}) \cdot \widetilde{O}\big(n + n^{3/4}\sqrt{L_1/\epsilon_H}\big) + O(L_1 \Delta_F \epsilon^{-2} n^{-1/2}) \widetilde{O}(n)$$
$$= \widetilde{O}\big(\Delta_F n L_3 \epsilon_H^{-2} + \Delta_F n^{3/4} L_1^{1/2} L_3 \epsilon_H^{-5/2} + \Delta_F n^{1/2} L_1 \epsilon^{-2}\big). \tag{B.10}$$

Using Markov inequality, with probability at least 2/3, we have

$$T_{\text{total}} = \widetilde{O}\big(\Delta_F n L_3 \epsilon_H^{-2} + \Delta_F n^{3/4} L_1^{1/2} L_3 \epsilon_H^{-5/2} + \Delta_F n^{1/2} L_1 \epsilon^{-2}\big).$$

Note that $|\mathcal{I}_1|\delta = |\mathcal{I}_1|/(72 \cdot L_3 \Delta_F \epsilon_H^{-2}) \leq 1/12$. By union bound, with probability at least $1 - 1/3 - 1/3 - |\mathcal{I}_1|\delta \geq 1/4$, SNVRG$^+$ + Neon2$^{\text{finite}}$ will find an $(\epsilon, \epsilon_H)$-second order stationary point within

$$\widetilde{O}\big(\Delta_F n L_3 \epsilon_H^{-2} + \Delta_F n^{3/4} L_1^{1/2} L_3 \epsilon_H^{-5/2} + \Delta_F n^{1/2} L_1 \epsilon^{-2}\big) \tag{B.11}$$

stochastic gradient computations. $\square$

## B.2 Proof of Theorem 6.4

The following lemma shows that the negative curvature descent step (Line 12) of Algorithm 3 achieves more function value decrease under third-order smoothness assumption. The proof can be found in Lemma 4.6 of Yu et al. (2017a).

**Lemma B.2.** (Yu et al., 2017a) Suppose that $F(\mathbf{x}) = \mathbb{E}_{\xi \in \mathcal{D}} F(\mathbf{x}; \xi)$, each $F(\mathbf{x}; \xi)$ is $L_1$-smooth, $L_2$-Hessian Lipschitz continuous and $F(\mathbf{x})$ is $L_3$-third-order smooth. Let $\epsilon_H \in (0,1)$ and $\eta = \sqrt{3\epsilon_H/L_3}$. Suppose that $\lambda_{\min}(\nabla^2 F(\mathbf{z}_{u-1})) < -\epsilon_H$ and that at the $u$-th iteration Algorithm 3 executes the Neon2$^{\text{online}}$ algorithm (Line 7). Then with probability $1 - \delta$ it holds that

$$\mathbb{E}_\zeta \big[ F(\mathbf{z}_u) - F(\mathbf{z}_{u-1}) \big] \leq -\epsilon_H^2/(6L_3). \tag{B.12}$$

In addition, Neon2$^{\text{online}}$ takes $O(L_1^2/\epsilon_H^2 \log^2(d/\delta))$ stochastic gradient computations.

*Proof of Theorem 6.4.* Denote $\delta_0 = 1/(2500 C_1 \rho \Delta_F L_1 B_0^{-1/2} \epsilon^{-2})$, then by (6.2) we have $B_0 > 32\sigma^2/\epsilon^2 (1 + \log^{1/2}(1/\delta_0))^2$. Let $\mathcal{I} = \{1, \ldots, U\}$ be the index set of all iterations. We use $\mathcal{I}_1$ and $\mathcal{I}_2$ to represent the index set of iterates where $\mathbf{z}_u$ is obtained from Neon2$^{\text{online}}$ and One-epoch-SNVRG$^+$ respectively. Obviously $U = |\mathcal{I}_1| + |\mathcal{I}_2|$ and we need to upper bound $|\mathcal{I}_1|$ and $|\mathcal{I}_2|$. From Lemma B.2, we have with probability at least $1 - \delta$ that

$$\mathbb{E}\big[ F(\mathbf{z}_u) - F(\mathbf{z}_{u-1}) \big] \leq -\epsilon_H^2/(6L_3), \qquad \text{for } u \in \mathcal{I}_1. \tag{B.13}$$

By Corollary A.3, we have

$$\mathbb{E}\|\nabla F(\mathbf{z}_u)\|_2^2 \leq C_1 \bigg( \frac{\rho L_1}{B_0^{1/2}} \cdot \mathbb{E}\big[ F(\mathbf{z}_{u-1}) - F(\mathbf{z}_u) \big] + \frac{\sigma^2}{B_0} \bigg), \qquad \text{for } u \in \mathcal{I}_2, \tag{B.14}$$

where $C_1 = 200$. We decompose $\mathcal{I}_2$ into two disjoint sets $\mathcal{I}_2 = \mathcal{I}_2^1 \cup \mathcal{I}_2^2$, where $\mathcal{I}_2^1 = \{u \in \mathcal{I}_2 : \|\mathbf{g}_u\|_2 > \epsilon/2\}$ and $\mathcal{I}_2^1 = \{u \in \mathcal{I}_2 : \|\mathbf{g}_u\|_2 \leq \epsilon/2\}$. (B.14) leads to the following inequalities:

$$\mathbb{E}\big[ F(\mathbf{z}_u) - F(\mathbf{z}_{u-1}) \big] \leq -\frac{B_0^{1/2}}{C_1 \rho L_1} \mathbb{E}\|\nabla F(\mathbf{z}_u)\|_2^2 + \frac{\sigma^2}{\rho L_1 B_0^{1/2}}, \qquad \text{for } u \in \mathcal{I}_2^1, \tag{B.15}$$

$$\mathbb{E}\big[ F(\mathbf{z}_u) - F(\mathbf{z}_{u-1}) \big] \leq \frac{\sigma^2}{\rho L_1 B_0^{1/2}}, \qquad \text{for } u \in \mathcal{I}_2^2. \tag{B.16}$$

Summing up (B.13) over $u \in \mathcal{I}_1$, (B.15) over $u \in \mathcal{I}_2^1$ and (B.16) over $u \in \mathcal{I}_2^2$, we have

$$\mathbb{E}\bigg[\sum_{u \in \mathcal{I}} F(\mathbf{z}_{u-1}) - F(\mathbf{z}_u)\bigg] \geq |\mathcal{I}_1| \cdot \frac{\epsilon_H^2}{6L_3} + \frac{B_0^{1/2}}{C_1 \rho L_1} \sum_{u \in \mathcal{I}_2^1} \mathbb{E}\|\nabla F(\mathbf{z}_u)\|_2^2 - \sum_{u \in \mathcal{I}_2^1} \frac{\sigma^2}{\rho L_1 B_0^{1/2}} - \sum_{u \in \mathcal{I}_2^2} \frac{\sigma^2}{\rho L_1 B_0^{1/2}}. \tag{B.17}$$

For any $u \in \mathcal{I}_2^2$, we have $\|\mathbf{g}_u\|_2 \leq \epsilon/2$, then algorithm will execute Neon2$^{\text{online}}$ at $u$-th iteration, which implies $|\mathcal{I}_2^2| \leq |\mathcal{I}_1|$. Combining this with (B.17) and by the definition of $\Delta_F$, with probability at least $1 - |\mathcal{I}_1|\delta$, we have

$$\frac{|\mathcal{I}_1|\epsilon_H^2}{6L_3} + \frac{B_0^{1/2}}{C_1 \rho L_1} \sum_{u \in \mathcal{I}_2^1} \mathbb{E}\|\nabla F(\mathbf{z}_u)\|_2^2 \leq \Delta_F + (|\mathcal{I}_1| + |\mathcal{I}_2^1|) \frac{\sigma^2}{\rho L_1 B_0^{1/2}}. \tag{B.18}$$

Applying Markov inequality, yields with probability at least $2/3$ that

$$\frac{|\mathcal{I}_1|\epsilon_H^2}{6L_3} + \frac{B_0^{1/2}}{C_1\rho L_1}\sum_{u\in\mathcal{I}_2^1}\|\nabla F(\mathbf{z}_u)\|_2^2 \leq 3\left(\Delta_F + (|\mathcal{I}_1|+|\mathcal{I}_2^1|)\frac{\sigma^2}{\rho L_1 B_0^{1/2}}\right). \tag{B.19}$$

By Lemma A.5, if $B_0 \geq 32\sigma^2/\epsilon^2(1+\log^{1/2}(1/\delta_0))^2$, then for any $u \in \mathcal{I}_2^1$, with probability at least $1-\delta_0$, we have $\|\nabla F(\mathbf{z}_u)\|_2 > \epsilon/4$. Applying union bound, we have with probability at least $1-|\mathcal{I}_1|\delta - 1/3 - |\mathcal{I}_2^1|\delta_0$ it holds that

$$\frac{|\mathcal{I}_1|\epsilon_H^2}{6L_3} + \frac{|\mathcal{I}_2^1|B_0^{1/2}\epsilon^2}{16C_1\rho L_1} \leq 3\Delta_F + \frac{3|\mathcal{I}_2^1|\sigma^2}{\rho L_1 B_0^{1/2}} + \frac{3|\mathcal{I}_1|\sigma^2}{\rho L_1 B_0^{1/2}}. \tag{B.20}$$

By (6.2) we have $B_0 > 96C_1\sigma^2\epsilon^{-2}$, which indicates

$$\frac{3|\mathcal{I}_2^1|\sigma^2}{\rho L_1 B_0^{1/2}} \leq \frac{|\mathcal{I}_2^1|B_0^{1/2}\epsilon^2}{32C_1\rho L_1}. \tag{B.21}$$

By the choice of $\rho$ we have $\rho = \max\{36\sigma^2 L_1^{-1}L_3\epsilon_H^{-2}B_0^{-1/2}, 6\} \geq 36\sigma^2 L_1^{-1}L_3\epsilon_H^{-2}B_0^{-1/2}$, which indicates

$$\frac{3|\mathcal{I}_1|\sigma^2}{\rho L_1 B_0^{1/2}} \leq \frac{|\mathcal{I}_1|\epsilon_H^2}{12L_3}. \tag{B.22}$$

Plugging (B.21), (B.22) into (B.20), then with probability $1-|\mathcal{I}_1|\delta - 1/3 - |\mathcal{I}_2^1|\delta_0$, we have

$$\frac{|\mathcal{I}_1|\epsilon_H^2}{12L_3} + \frac{|\mathcal{I}_2^1|B_0^{1/2}\epsilon^2}{32C_1\rho L_1} \leq 3\Delta_F,$$

which immediately implies

$$|\mathcal{I}_1| \leq 36\Delta_F L_3 \epsilon_H^{-2} = O(\Delta_F L_3 \epsilon_H^{-2}), \tag{B.23}$$

and

$$|\mathcal{I}_2^1| \leq 96C_1\rho\Delta_F L_1 B_0^{-1/2}\epsilon^{-2} = \widetilde{O}(\Delta_F \sigma^{-1}L_1\epsilon^{-1} + \Delta_F L_3 \epsilon_H^{-2}). \tag{B.24}$$

Total number of iteration is $U = |\mathcal{I}_1|+|\mathcal{I}_2| \leq 2|\mathcal{I}_1|+|\mathcal{I}_2^1| \leq 72\Delta_F L_3 \epsilon_H^{-2} + 96C_1\rho\Delta_F L_1 B_0^{-1/2}\epsilon^{-2}$. Now we calculate the gradient complexity of Algorithm 3. By Lemma B.2 one single call of Neon2$^{\text{online}}$ needs $O(L_1^2/\epsilon_H^2\log^2(d/\delta))$ stochastic gradient computations and by Corollary A.3 one single call of One-epoch-SNVRG$^+$ needs $20B_0\log^3 B_0 = \widetilde{O}(\sigma^2/\epsilon^2)$ stochastic gradient computations. In addition, we need to compute $\mathbf{g}_u$ at each iteration, which takes $\widetilde{O}(\sigma^2/\epsilon^2)$ stochastic gradient computations. The expected total amount of stochastic gradient computations $\mathbb{E}T_{\text{total}}$ is

$$\begin{aligned}\mathbb{E}T_{\text{total}} &= |\mathcal{I}_1|\cdot O(L_1^2/\epsilon_H^2\log^2(d/\delta)) + |\mathcal{I}_2|\cdot \widetilde{O}(\sigma^2/\epsilon^2) + |\mathcal{I}|\cdot \widetilde{O}(\sigma^2/\epsilon^2)\\ &= |\mathcal{I}_1|\cdot \widetilde{O}(L_1^2/\epsilon_H^2) + (|\mathcal{I}_2^1|+|\mathcal{I}_2^2|)\cdot \widetilde{O}(\sigma^2/\epsilon^2) + |\mathcal{I}|\cdot \widetilde{O}(\sigma^2/\epsilon^2)\\ &= |\mathcal{I}_1|\cdot \widetilde{O}(L_1^2/\epsilon_H^2) + (|\mathcal{I}_2^1|+|\mathcal{I}_1|)\cdot \widetilde{O}(\sigma^2/\epsilon^2)\\ &= \widetilde{O}(\Delta_F L_1^2 L_3\epsilon_H^{-4} + \Delta_F\sigma^2 L_3 \epsilon_H^{-2}\epsilon^{-2} + \Delta_F\sigma L_1\epsilon^{-3}).\end{aligned} \tag{B.25}$$

Applying Markov's inequality, with probability at least 2/3, we have

$$T_{\text{total}} = \widetilde{O}(\Delta_F L_1^2 L_3 \epsilon_H^{-4} + \Delta_F \sigma^2 L_3 \epsilon_H^{-2} \epsilon^{-2} + \Delta_F \sigma L_1 \epsilon^{-3}).$$

Note that $|\mathcal{I}_1|\delta = |\mathcal{I}_1|/(1000\Delta_F L_3 \epsilon_H^{-2}) \leq 1/24$ and $|\mathcal{I}_2^1|\delta_0 = |\mathcal{I}_2^1|/(2500 C_1 \rho \Delta_F L_1 B_0^{-1/2} \epsilon^{-2}) < 1/24$. Therefore, with probability at least $1 - |\mathcal{I}_1|\delta - 1/3 - |\mathcal{I}_2^1|\delta_0 - 1/3 \geq 1/4$, Algorithm 3 can find an $(\epsilon, \epsilon_H)$-second-order stationary point with

$$\widetilde{O}(\Delta_F L_1^2 L_3 \epsilon_H^{-4} + \Delta_F \sigma^2 L_3 \epsilon_H^{-2} \epsilon^{-2} + \Delta_F \sigma L_1 \epsilon^{-3}) \tag{B.26}$$

stochastic gradient computations. $\square$

## C Proof of Lemma 5.1

Now we prove our key lemma, i.e., Lemma 5.1. We need the following supporting lemmas. The first lemma shows that with any chosen epoch length $T$, the summation of expectation of the square of gradient norm $\sum_{j=0}^{T-1} \mathbb{E}\|\nabla F(\mathbf{x}_j)\|_2^2$ can be bounded.

**Lemma C.1.** Suppose $T > 0$ is the amount of epochs which is defined in Algorithm 1. We fix $T$ and suppose that $T > 1$. If the step size and batch size parameters in Algorithm 1 satisfy $M \geq 6L$ and $B_l \geq 6^{K-l+1}(\prod_{s=l}^{K} T_s)^2$ for any $1 \leq l \leq K$, then the output of Algorithm 1 satisfies

$$\sum_{j=0}^{T-1} \mathbb{E}\|\nabla F(\mathbf{x}_j)\|_2^2 \leq C\bigg(M\mathbb{E}\big[F(\mathbf{x}_0) - F(\mathbf{x}_T)\big] + \frac{2\sigma^2 T}{B_0} \cdot \mathbb{1}\{B_0 < n\}\bigg), \tag{C.1}$$

where $C = 100$.

The next lemma is about geometric distribution, which helps us to only consider the last term of a sequence rather than consider the whole sequence under the geometric distribution assumption.

**Lemma C.2.** Suppose that $G \sim \text{Geom}(p)$, where $\mathbb{P}(G = k) = p(1-p)^k, k \geq 0$. Let $a(j), b(j)$ be two series and $b(0) \geq 0$. If for any $k \geq 1$, it holds that $\sum_{j=0}^{k-1} a(j) \leq b(k)$, then we have

$$\frac{1-p}{p}\mathbb{E}_G a(G) \leq \mathbb{E}_G b(G). \tag{C.2}$$

*Proof of Lemma 5.1.* We can easily check that the choice of $M, \{T_l\}, \{B_l\}$ in Lemma 5.1 satisfies the assumption of Lemma C.1. By Algorithm 1, we have $T \sim \text{Geom}(p)$ where $p = 1/(1 + \prod_{j=1}^{K} T_j)$. Let

$$a(j) = \mathbb{E}\|\nabla F(\mathbf{x}_j)\|_2^2, \ b(j) = C\bigg(M\mathbb{E}\big[F(\mathbf{x}_0) - F(\mathbf{x}_j)\big] + \frac{\sigma^2 j}{B_0} \cdot \mathbb{1}\{B_0 < n\}\bigg). \tag{C.3}$$

Then by Lemma C.1, for any $T \geq 1$, we have $\sum_{j=0}^{T-1} a(j) \leq b(T)$ and $b(0) = 0$. Thus, by Lemma C.2, we have

$$\frac{1-p}{p}\mathbb{E}_T \mathbb{E}\|\nabla F(\mathbf{x}_T)\|_2^2 \leq C\bigg(M\mathbb{E}_T \mathbb{E}\big[F(\mathbf{x}_0) - F(\mathbf{x}_T)\big] + \frac{2\sigma^2 \mathbb{E}_T T}{B_0} \cdot \mathbb{1}\{B_0 < n\}\bigg). \tag{C.4}$$

Since $\mathbb{E}_T T = (1-p)/p = \prod_{j=1}^{K} T_j = B_0^{1/2}$, then we have

$$\mathbb{E}\|\nabla F(\mathbf{x}_T)\|_2^2 \leq C\bigg(\frac{M}{\prod_{j=1}^{K} T_j}\mathbb{E}\big[F(\mathbf{x}_0) - F(\mathbf{x}_T)\big] + \frac{2\sigma^2}{B_0}\cdot \mathbb{1}\{B_0 < n\}\bigg)$$

$$= C\bigg(\frac{M}{B_0^{1/2}}\mathbb{E}\big[F(\mathbf{x}_0) - F(\mathbf{x}_T)\big] + \frac{2\sigma^2}{B_0}\cdot \mathbb{1}\{B_0 < n\}\bigg),$$

which immediately implies (5.1).

Finally we consider how many stochastic gradient computations for us to run One-epoch-SNVRG$^+$ once. According to the update of reference gradients in Algorithm 1, for $\mathbf{g}_t^{(l)}$, we need to update it when $0 = (t \mod \prod_{j=l+1}^{K} T_j)$, and thus we need to sample $\mathbf{g}_t^{(l)}$ for $T/\prod_{j=l+1}^{K} T_j$ times. We need $B_0$ stochastic gradient computations to update $\mathbf{g}_t^{(0)}$ and $2B_l$ stochastic gradient computations for $\mathbf{g}_t^{(l)}$ (Lines 18 and 21 in Algorithm 1 respectively). If we use $\mathcal{T}$ to represent the total number of stochastic gradient computations, then based on above arguments, we have

$$\mathbb{E}\mathcal{T} = B_0 \cdot \frac{\mathbb{E}T}{\prod_{j=1}^{K} T_j} + 2\sum_{l=1}^{K} B_l \cdot \frac{\mathbb{E}T}{\prod_{j=l+1}^{K} T_j}$$

$$= B_0 + 2\sum_{l=1}^{K} B_l \cdot \prod_{j=1}^{l} T_j \tag{C.5}$$

Now we calculate $\mathbb{E}\mathcal{T}$ under the parameter choice of Lemma 5.1. Note that we can easily verify the following results:

$$\prod_{j=1}^{l} T_j = 2^{2^{l-1}} = B_0^{\frac{2^l}{2^{K+1}}}, \quad B_1 \cdot \prod_{j=1}^{1} T_j = 2 \times 6^K B_0, \quad B_l \cdot \prod_{j=1}^{l} T_j = 6^{K-l+1} B_0. \tag{C.6}$$

Submit (C.6) into (C.5) yields the following results:

$$\mathbb{E}\mathcal{T} = B_0 + 2\bigg(2 \times 6^K B_0 + \sum_{l=2}^{K} 6^{K-l+1} B_0\bigg)$$

$$< B_0 + 6 \times 6^K B_0$$

$$= B_0 + 6 \times 6^{\log\log B_0} B_0$$

$$< B_0 + 6 B_0 \log^3 B_0. \tag{C.7}$$

Therefore, the expectation of total gradient complexity $\mathbb{E}\mathcal{T}$ is bounded by the following:

$$\mathbb{E}\mathcal{T} \leq B_0 + 6 B_0 \log^3 B_0 \leq 7 B_0 \log^3 B_0, \tag{C.8}$$

which completes the proof. $\square$

## D  Proof of Supporting Lemmas

In this section, we prove the supporting lemmas used in the proof of Lemma 5.1.

## D.1 Proof of Lemma C.2

*Proof.* Since $G$ follows geometric distribution $\text{Geom}(p)$, we first obtain

$$\mathbb{E}_G b(G) = \sum_{k=0}^{\infty} p(1-p)^k b(k). \tag{D.1}$$

Note that for any $k \geq 1$, it holds that $\sum_{j=0}^{k-1} a(j) \leq b(k)$. Plugging the inequality into (D.1) yields

$$\begin{aligned}
\mathbb{E}_G b(G) &\geq \sum_{k=1}^{\infty} p(1-p)^k \sum_{j=0}^{k-1} a(j) \\
&= \sum_{j=0}^{\infty} a(j) \sum_{k=j+1}^{\infty} p(1-p)^k \\
&= \sum_{j=0}^{\infty} a(j)(1-p)^{j+1} \\
&= \frac{1-p}{p} \sum_{j=0}^{\infty} a(j) p(1-p)^j \\
&= \frac{1-p}{p} \mathbb{E}_G a(G),
\end{aligned} \tag{D.2}$$

which completes the proof. $\square$

## D.2 Proof of Lemma C.1

Now we provide the proof of Lemma C.1 which is inspired by the same high level idea as the proof of Lemma A.1 in Zhou et al. (2018a). Nevertheless, our One-epoch-SNVRG$^+$ algorithm is different from One-epoch-SNVRG in Zhou et al. (2018a) from several aspects and so will be the proofs of the supporting lemmas. Let $M, \{T_i\}, \{B_i\}, B_0$ be the variables defined in One-epoch-SNVRG$^+$ (Algorithm 1) and $\{\mathbf{x}_j\}$ be the iterates generated by Algorithm 1. We define a filtration $\mathcal{F}_t = \sigma(\mathbf{x}_0, \ldots, \mathbf{x}_t)$ and $\{\mathbf{x}_t^{(l)}\}, \{\mathbf{g}_t^{(l)}\}$ are the reference points and reference gradients used in Algorithm 1. By the updating rules of $\{\mathbf{x}_t^{(l)}\}$ in Algorithm 1, we have

$$\mathbf{x}_t^{(l)} = \mathbf{x}_{t^l}, \ t^l = \left\lfloor \frac{t}{\prod_{j=l+1}^K T_j} \right\rfloor \prod_{j=l+1}^K T_j. \tag{D.3}$$

We further define $\mathbf{v}_t^{(l)}$ as

$$\mathbf{v}_t^{(l)} := \sum_{j=0}^{l} \mathbf{g}_t^{(j)}, \quad \text{for } 0 \leq l \leq K. \tag{D.4}$$

**Definition D.1.** (Zhou et al., 2018a) Constant series $\{c_j^{(s)}\}$ is defined as follows. For each $s$, we define $c_{T_s}^{(s)}$ as

$$c_{T_s}^{(s)} = \frac{M}{6^{K-s+1} \prod_{l=s}^K T_l}. \tag{D.5}$$

When $0 \leq j < T_s$, we define $c_j^{(s)}$ by induction:

$$c_j^{(s)} = \left(1 + \frac{1}{T_s}\right) c_{j+1}^{(s)} + \frac{3L^2}{M} \cdot \frac{\prod_{l=s+1}^{K} T_l}{B_s}. \tag{D.6}$$

The following lemma characterizes the property of the constant series defined above.

**Lemma D.2.** (Zhou et al., 2018a) If $B_s \geq 6^{K-s+1}(\prod_{l=s}^{K} T_l)^2, T_l \geq 1$ and $M \geq 6L$, then it holds that

$$c_j^{(s-1)} \cdot (1 + T_{s-1}) < c_{T_s}^{(s)}, \qquad \text{for } 2 \leq s \leq K, 0 \leq j \leq T_{s-1}, \tag{D.7}$$

and

$$c_j^{(K)} \cdot (1 + T_K) < M, \qquad \text{for } 0 \leq j \leq T_K. \tag{D.8}$$

We also need the following two technical lemmas in our proof.

**Lemma D.3.** For any $p, s$, where $1 \leq s \leq K$, $p \cdot \prod_{j=s}^{K} T_j < T$ and $q \prod_{j=1}^{K} T_j \leq p \cdot \prod_{j=s}^{K} T_j < (p+1) \cdot \prod_{j=s}^{K} T_j \leq (q+1) \prod_{j=1}^{K} T_j$, we define

$$\text{start} = p \cdot \prod_{j=s}^{K} T_j, \quad \text{end} = \min\left\{\text{start} + \prod_{j=s}^{K} T_j, T\right\}$$

for simplification. Then we have the following results:

$$\mathbb{E}\left[\sum_{j=\text{start}}^{\text{end}-1} \frac{\|\nabla F(\mathbf{x}_j)\|_2^2}{100M} + F(\mathbf{x}_{\text{end}}) + c_{T_s}^{(s)} \cdot \|\mathbf{x}_{\text{end}} - \mathbf{x}_{\text{start}}\|_2^2 \Big| \mathcal{F}_{\text{start}}\right]$$
$$\leq F(\mathbf{x}_{\text{start}}) + \frac{2}{M} \cdot \mathbb{E}\left[\|\nabla F(\mathbf{x}_{\text{start}}) - \mathbf{v}_{\text{start}}\|_2^2 \Big| \mathcal{F}_{\text{start}}\right] \cdot (\text{end} - \text{start}).$$

**Lemma D.4.** (Lei et al., 2017) If $\mathbf{a}_i$ are vectors satisfying $\sum_{i=1}^{N} \mathbf{a}_i = 0$ and $\mathcal{J}$ is a uniform random subset of $\{1, \ldots, N\}$ with size $m$, then

$$\mathbb{E}_{\mathcal{J}} \left\| \frac{1}{m} \sum_{j \in \mathcal{J}} \mathbf{a}_j \right\|_2^2 \leq \frac{\mathbb{1}(|\mathcal{J}| < N)}{m} \mathbb{E}_j \|\mathbf{a}_j\|_2^2.$$

*Proof of Lemma C.1.* For any $q$ which satisfies that $q \cdot \prod_{j=1}^{K} T_j < T$, we denote

$$\text{start}^{(q)} = q \cdot \prod_{j=1}^{K} T_j, \quad \text{end}^{(q)} = \min\left\{\text{start}^{(q)} + \prod_{j=1}^{K} T_j, T\right\},$$

then we have

$$\sum_{j=\text{start}^{(q)}}^{\text{end}^{(q)}-1} \frac{\mathbb{E}\|\nabla F(\mathbf{x}_j)\|_2^2}{100M} + \mathbb{E}\left[F(\mathbf{x}_{\text{end}^{(q)}})\right]$$

$$\leq \sum_{j=\text{start}^{(q)}}^{\text{end}^{(q)}-1} \frac{\mathbb{E}\|\nabla F(\mathbf{x}_j)\|_2^2}{100M} + \mathbb{E}\big[F(\mathbf{x}_{\text{end}^{(q)}}) + c_{T_1}^{(1)} \cdot \|\mathbf{x}_{\text{end}^{(q)}} - \mathbf{x}_{\text{start}^{(q)}}\|_2^2\big]$$

$$\leq \mathbb{E}\big[F(\mathbf{x}_{\text{start}^{(q)}})\big] + \frac{2}{M} \cdot \mathbb{E}\|\nabla F(\mathbf{x}_{\text{start}^{(q)}}) - \mathbf{g}_{\text{start}^{(q)}}^{(0)}\|_2^2 \cdot (\text{end}^{(q)} - \text{start}^{(q)}), \tag{D.9}$$

where the second inequality comes from Lemma D.3, with $s=1, p=q$. Moreover we have

$$\mathbb{E}\|\nabla F(\mathbf{x}_{\text{start}^{(q)}}) - \mathbf{g}_{\text{start}^{(q)}}^{(0)}\|_2^2 = \mathbb{E}\bigg\|\frac{1}{B_0}\sum_{i\in I}\big[\nabla f_i(\mathbf{x}_{\text{start}^{(q)}}) - \nabla F(\mathbf{x}_{\text{start}^{(q)}})\big]\bigg\|_2^2$$

$$\leq \mathbb{1}\{B_0 < n\} \cdot \frac{1}{B_0}\mathbb{E}\|\nabla f_i(\mathbf{x}_{\text{start}^{(q)}}) - \nabla F(\mathbf{x}_{\text{start}^{(q)}})\|_2^2 \tag{D.10}$$

$$\leq \mathbb{1}\{B_0 < n\} \cdot \frac{\sigma^2}{B_0}, \tag{D.11}$$

where (D.10) holds because of Lemma D.4 and (D.11) holds because by the sub-Gaussian stochastic gradient assumption we made on $F$, we have $\mathbb{E}\|\nabla f_i(\mathbf{x}_{\text{start}^{(q)}}) - \nabla F(\mathbf{x}_{\text{start}^{(q)}})\|_2^2 \leq \sigma^2$. Plugging (D.11) into (D.9), we obtain

$$\sum_{j=\text{start}^{(q)}}^{\text{end}^{(q)}-1} \mathbb{E}\|\nabla F(\mathbf{x}_j)\|_2^2 \leq C\bigg(M\mathbb{E}\big[F(\mathbf{x}_{\text{start}^{(q)}}) - F(\mathbf{x}_{\text{end}^{(q)}})\big] + \frac{2(\text{end}^{(q)} - \text{start}^{(q)})\sigma^2}{B_0} \cdot \mathbb{1}\{B_0 < n\}\bigg), \tag{D.12}$$

where $C = 100$. Suppose that $q^* = \max\{q : \text{start}^{(q)} < T\}$. Telescoping (D.12) for $q = 0$ to $q^*$, we have

$$\sum_{q=0}^{q^*}\sum_{j=\text{start}^{(q)}}^{\text{end}^{(q)}-1} \mathbb{E}\|\nabla F(\mathbf{x}_j)\|_2^2$$

$$\leq C\bigg(M\sum_{q=0}^{q^*}\mathbb{E}\big[F(\mathbf{x}_{\text{start}^{(q)}}) - F(\mathbf{x}_{\text{end}^{(q)}})\big] + \frac{2\sum_{q=0}^{q^*}(\text{end}^{(q)} - \text{start}^{(q)})\sigma^2}{B_0} \cdot \mathbb{1}\{B_0 < n\}\bigg) \tag{D.13}$$

Note that for $1 \leq q \leq q^*$, we have $\text{start}^{(0)} = 0, \text{start}^{(q)} = \text{end}^{(q-1)}$ and $\text{end}^{(q^*)} = T$. Thus, (D.13) equals

$$\sum_{j=0}^{T-1}\mathbb{E}\|\nabla F(\mathbf{x}_j)\|_2^2 \leq C\bigg(M\mathbb{E}\big[F(\mathbf{x}_0) - F(\mathbf{x}_T)\big] + \frac{2\sigma^2 T}{B_0} \cdot \mathbb{1}\{B_0 < n\}\bigg),$$

which completes the proof. $\square$

# E  Proof of Technical Lemmas

In this section, we provide the proofs of technical lemmas used in the proof of Lemma C.1.

## E.1 Proof of Lemma D.3

Let $M, \{T_l\}, \{B_l\}, B_0$ be the parameters in Algorithm 1, $\{\mathbf{x}_t\}$ be the iterates generated by Algorithm 1, $T$ be the amount of epochs in Algorithm 1 and $\{\mathbf{x}_t^{(l)}\}, \{\mathbf{g}_t^{(l)}\}$ be the reference points and reference gradients defined in Algorithm 1. Let $\mathbf{v}_t^{(l)}, \mathcal{F}_t$ and $c_j^{(s)}$ be defined as in Appendix D. To prove Lemma D.3, we need the following lemmas.

**Lemma E.1.** (Zhou et al., 2018a) Let $\mathbf{v}_t^{(l)}$ be defined as in (D.4). Let $p, s$ satisfy $q \prod_{j=1}^K T_j \leq p \cdot \prod_{j=s+1}^K T_j < (p+1) \cdot \prod_{j=s+1}^K T_j < (q+1) \prod_{j=1}^K T_j$ for some $q$. For any $t, t'$ satisfying $p \cdot \prod_{j=s+1}^K T_j \leq t < t' < (p+1) \cdot \prod_{j=s+1}^K T_j$, it holds that

$$\mathbf{x}_t^{(s)} = \mathbf{x}_{t'}^{(s)} = \mathbf{x}_{p \prod_{j=s+1}^K T_j}, \tag{E.1}$$

$$\mathbf{g}_t^{(s')} = \mathbf{g}_{t'}^{(s')}, \qquad \text{for any } s' \text{ that satisfies } 0 \leq s' \leq s, \tag{E.2}$$

$$\mathbf{v}_t^{(s)} = \mathbf{v}_{t'}^{(s)} = \mathbf{v}_{p \prod_{j=s+1}^K T_j}. \tag{E.3}$$

The following lemma is a special case of Lemma D.3 when $s = K$:

**Lemma E.2.** Suppose $p$ satisfies $q \prod_{i=1}^K T_i \leq pT_K < (p+1)T_K \leq (q+1) \prod_{i=1}^K T_i$ for some $q$ and $pT_K < T$. For simplification, we denote

$$\text{start} = pT_K, \text{end} = \min\{(p+1)T_K, T\}. \tag{E.4}$$

If $M > L$, then we have

$$\mathbb{E}\left[F(\mathbf{x}_{\text{end}}) + c_{T_K}^{(K)} \cdot \|\mathbf{x}_{\text{end}} - \mathbf{x}_{\text{start}}\|_2^2 + \sum_{j=\text{start}}^{\text{end}-1} \frac{\|\nabla F(\mathbf{x}_j)\|_2^2}{100M}\bigg|\mathcal{F}_{\text{start}}\right]$$
$$\leq F(\mathbf{x}_{\text{start}}) + \frac{2}{M} \cdot \mathbb{E}\left[\|\nabla F(\mathbf{x}_{\text{start}}) - \mathbf{v}_{\text{start}}\|_2^2 | \mathcal{F}_{\text{start}}\right] \cdot (\text{end} - \text{start}). \tag{E.5}$$

The following lemma provides an upper bound of $\mathbb{E}\left[\|\nabla F(\mathbf{x}_t^{(l)}) - \mathbf{v}_t^{(l)}\|_2^2\right]$.

**Lemma E.3** (Zhou et al. (2018a)). Let $t^l$ be as defined in (D.3), then we have $\mathbf{x}_t^{(l)} = \mathbf{x}_{t^l}$, and

$$\mathbb{E}\left[\|\nabla F(\mathbf{x}_t^{(l)}) - \mathbf{v}_t^{(l)}\|_2^2 | \mathcal{F}_{t^l}\right] \leq \frac{L^2}{B_l}\|\mathbf{x}_t^{(l)} - \mathbf{x}_t^{(l-1)}\|_2^2 + \|\nabla F(\mathbf{x}_t^{(l-1)}) - \mathbf{v}_t^{(l-1)}\|_2^2.$$

*Proof of Lemma D.3.* We use mathematical induction to prove that Lemma D.3 holds for any $1 \leq s \leq K$. When $s = K$, we have the result hold because of Lemma E.2. Suppose that for $s+1$, Lemma D.3 holds for any $p'$ which satisfies $p' \prod_{j=s+1}^K T_j < T$ and $q \prod_{j=1}^K T_j \leq p' \prod_{j=s+1}^K T_j < (p'+1) \prod_{j=s+1}^K T_j \leq (q+1) \prod_{j=1}^K T_j$. We need to prove Lemma D.3 still holds for $s$ and $p$, where $p$ satisfies $p \prod_{j=s+1}^K T_j < T$ and $q \prod_{j=1}^K T_j \leq p \prod_{j=s}^K T_j < (p+1) \prod_{j=s}^K T_j \leq (q+1) \prod_{j=1}^K T_j$. We choose $p' = pT_s + u$ which satisfies that $p' \prod_{j=s+1}^K T_j < T$, and we set indices $\text{start}_u$ and $\text{end}_u$ as

$$\text{start}_u = p' \prod_{j=s+1}^K T_j, \qquad \text{end}_u = \min\left\{\text{start}_u + \prod_{j=s+1}^K T_j, T\right\}.$$

Then we have

$$\mathbb{E}\bigg[\sum_{j=\text{start}_u}^{\text{end}_u-1}\frac{\|\nabla F(\mathbf{x}_j)\|_2^2}{100M} + F(\mathbf{x}_{\text{end}_u}) + c_{T_{s+1}}^{(s+1)} \cdot \|\mathbf{x}_{\text{end}_u} - \mathbf{x}_{\text{start}_u}\|_2^2 \Big| \mathcal{F}_{\text{start}_u}\bigg]$$
$$\leq F(\mathbf{x}_{\text{start}_u}) + \frac{2}{M} \cdot \mathbb{E}\big[\|\nabla F(\mathbf{x}_{\text{start}_u}) - \mathbf{v}_{\text{start}_u}\|_2^2 \big| \mathcal{F}_{\text{start}_u}\big] \cdot (\text{end}_u - \text{start}_u), \quad (\text{E.6})$$

where the last inequality holds because of the induction hypothesis that Lemma D.3 holds for $s+1$ and $p'$. Note that we have $\mathbf{x}_{\text{start}_u} = \mathbf{x}_{\text{start}_u}^{(s)}$ from Proposition E.1, which implies

$$\mathbb{E}\big[\|\nabla F(\mathbf{x}_{\text{start}_u}) - \mathbf{v}_{\text{start}_u}\|_2^2 \big| \mathcal{F}_{\text{start}_u}\big] = \mathbb{E}\big[\|\nabla F(\mathbf{x}_{\text{start}_u}^{(s)}) - \mathbf{v}_{\text{start}_u}^{(s)}\|_2^2 \big| \mathcal{F}_{\text{start}_u}\big]$$
$$\leq \frac{L^2}{B_s}\|\mathbf{x}_{\text{start}_u}^{(s)} - \mathbf{x}_{\text{start}_u}^{(s-1)}\|_2^2 + \|\nabla F(\mathbf{x}_{\text{start}_u}^{(s-1)}) - \mathbf{v}_{\text{start}_u}^{(s-1)}\|_2^2 \quad (\text{E.7})$$
$$= \frac{L^2}{B_s}\|\mathbf{x}_{\text{start}_u} - \mathbf{x}_{\text{start}}\|_2^2 + \|\nabla F(\mathbf{x}_{\text{start}}) - \mathbf{v}_{\text{start}}\|_2^2, \quad (\text{E.8})$$

where (E.7) holds because of Lemma E.3 and (E.8) holds due to Proposition E.1. Plugging (E.8) into (E.6) and taking expectation $\mathbb{E}[\cdot|\mathcal{F}_{\text{start}}]$ for (E.6) will yield

$$\mathbb{E}\bigg[\sum_{j=\text{start}_u}^{\text{end}_u-1}\frac{\|\nabla F(\mathbf{x}_j)\|_2^2}{100M} + F(\mathbf{x}_{\text{end}_u}) + c_{T_{s+1}}^{(s+1)}\|\mathbf{x}_{\text{end}_u} - \mathbf{x}_{\text{start}_u}\|_2^2 \Big| \mathcal{F}_{\text{start}}\bigg]$$
$$\leq \mathbb{E}\bigg[F(\mathbf{x}_{\text{start}_u}) + (\text{end}_u - \text{start}_u)\frac{2L^2}{MB_s}\|\mathbf{x}_{\text{start}_u} - \mathbf{x}_{\text{start}}\|_2^2$$
$$+ \frac{2(\text{end}_u - \text{start}_u)}{M}\|\nabla F(\mathbf{x}_{\text{start}}) - \mathbf{v}_{\text{start}}\|_2^2 \Big| \mathcal{F}_{\text{start}}\bigg]$$
$$\leq \mathbb{E}\bigg[F(\mathbf{x}_{\text{start}_u}) + \bigg(\prod_{j=s+1}^{K}T_j\bigg)\frac{2L^2}{MB_s}\|\mathbf{x}_{\text{start}_u} - \mathbf{x}_{\text{start}}\|_2^2$$
$$+ \frac{2(\text{end}_u - \text{start}_u)}{M}\|\nabla F(\mathbf{x}_{\text{start}}) - \mathbf{v}_{\text{start}}\|_2^2 \Big| \mathcal{F}_{\text{start}}\bigg]. \quad (\text{E.9})$$

We now give a bound of $\|\mathbf{x}_{\text{end}_u} - \mathbf{x}_{\text{start}}\|_2^2$:

$$\|\mathbf{x}_{\text{end}_u} - \mathbf{x}_{\text{start}}\|_2^2$$
$$= \|\mathbf{x}_{\text{start}_u} - \mathbf{x}_{\text{start}}\|_2^2 + \|\mathbf{x}_{\text{end}_u} - \mathbf{x}_{\text{start}_u}\|_2^2 + 2\langle\mathbf{x}_{\text{end}_u} - \mathbf{x}_{\text{start}_u}, \mathbf{x}_{\text{start}_u} - \mathbf{x}_{\text{start}}\rangle$$
$$\leq \|\mathbf{x}_{\text{start}_u} - \mathbf{x}_{\text{start}}\|_2^2 + \|\mathbf{x}_{\text{end}_u} - \mathbf{x}_{\text{start}_u}\|_2^2 + \frac{1}{T_s}\cdot\|\mathbf{x}_{\text{start}_u} - \mathbf{x}_{\text{start}}\|_2^2 + T_s\cdot\|\mathbf{x}_{\text{end}_u} - \mathbf{x}_{\text{start}_u}\|_2^2$$
$$(\text{E.10})$$
$$= \bigg(1 + \frac{1}{T_s}\bigg)\cdot\|\mathbf{x}_{\text{start}_u} - \mathbf{x}_{\text{start}}\|_2^2 + (1 + T_s)\cdot\|\mathbf{x}_{\text{end}_u} - \mathbf{x}_{\text{start}_u}\|_2^2, \quad (\text{E.11})$$

where (E.10) holds because of Young's inequality. Taking expectation $\mathbb{E}[\cdot|\mathcal{F}_{\text{start}}]$ over (E.11) and multiplying $c_{u+1}^{(s)}$ on both sides, we obtain

$$c_{u+1}^{(s)}\mathbb{E}\big[\|\mathbf{x}_{\text{end}_u} - \mathbf{x}_{\text{start}}\|_2^2 \big| \mathcal{F}_{\text{start}}\big] \leq c_{u+1}^{(s)}\bigg(1 + \frac{1}{T_s}\bigg)\mathbb{E}\big[\|\mathbf{x}_{\text{start}_u} - \mathbf{x}_{\text{start}}\|_2^2 \big| \mathcal{F}_{\text{start}}\big]$$

$$+ c_{u+1}^{(s)}(1 + T_s)\mathbb{E}\big[\|\mathbf{x}_{\text{end}_u} - \mathbf{x}_{\text{start}_u}\|_2^2 \big| \mathcal{F}_{\text{start}}\big]. \tag{E.12}$$

Adding up inequalities (E.12) and (E.9) together, we have

$$\mathbb{E}\bigg[\sum_{j=\text{start}_u}^{\text{end}_u-1} \frac{\|\nabla F(\mathbf{x}_j)\|_2^2}{100M} + F(\mathbf{x}_{\text{end}_u}) + c_{u+1}^{(s)}\|\mathbf{x}_{\text{end}_u} - \mathbf{x}_{\text{start}}\|_2^2 + c_{T_{s+1}}^{(s+1)}\|\mathbf{x}_{\text{end}_u} - \mathbf{x}_{\text{start}_u}\|_2^2 \bigg| \mathcal{F}_{\text{start}}\bigg]$$

$$\leq \mathbb{E}\bigg[F(\mathbf{x}_{\text{start}_u}) + \|\mathbf{x}_{\text{start}_u} - \mathbf{x}_{\text{start}}\|_2^2 \bigg[c_{u+1}^{(s)}\bigg(1 + \frac{1}{T_s}\bigg) + \frac{3L^2}{B_s M}\prod_{j=s+1}^{K} T_j\bigg] \bigg| \mathcal{F}_{\text{start}}\bigg]$$

$$+ \frac{2}{M}\mathbb{E}\big[\|\nabla F(\mathbf{x}_{\text{start}}) - \mathbf{v}_{\text{start}}\|_2^2 \big| \mathcal{F}_{\text{start}}\big](\text{end}_u - \text{start}_u)$$

$$+ c_{u+1}^{(s)}(1 + T_s)\mathbb{E}\big[\|\mathbf{x}_{\text{end}_u} - \mathbf{x}_{\text{start}_u}\|_2^2 \big| \mathcal{F}_{\text{start}}\big]$$

$$< \mathbb{E}\big[F(\mathbf{x}_{\text{start}_u}) + c_u^{(s)}\|\mathbf{x}_{\text{start}_u} - \mathbf{x}_{\text{start}}\|_2^2 \big| \mathcal{F}_{\text{start}}\big] + \frac{2}{M}\mathbb{E}\big[\|\nabla F(\mathbf{x}_{\text{start}}) - \mathbf{v}_{\text{start}}\|_2^2 \big| \mathcal{F}_{\text{start}}\big](\text{end}_u - \text{start}_u)$$

$$+ c_{T_{s+1}}^{(s+1)}\mathbb{E}\big[\|\mathbf{x}_{\text{end}_u} - \mathbf{x}_{\text{start}_u}\|_2^2 \big| \mathcal{F}_{\text{start}}\big], \tag{E.13}$$

where the last inequality holds due to the fact that $c_u^{(s)} = c_{u+1}^{(s)}(1 + 1/T_s) + 3L^2/(B_s M) \cdot \prod_{j=s+1}^{K} T_j$ by Definition D.1 and $c_{u+1}^{(s)} \cdot (1 + T_s) < c_{T_{s+1}}^{(s+1)}$ by Lemma D.2. Cancelling out the term $c_{T_{s+1}}^{(s+1)}\mathbb{E}\big[\|\mathbf{x}_{\text{end}_u} - \mathbf{x}_{\text{start}_u}\|_2^2 \big| \mathcal{F}_{\text{start}}\big]$ from both sides of (E.13), we get

$$\sum_{j=\text{start}_u}^{\text{end}_u-1} \mathbb{E}\bigg[\frac{\|\nabla F(\mathbf{x}_j)\|_2^2}{100M}\bigg|\mathcal{F}_{\text{start}}\bigg] + \mathbb{E}\big[F(\mathbf{x}_{\text{end}_u}) + c_{u+1}^{(s)} \cdot \|\mathbf{x}_{\text{end}_u} - \mathbf{x}_{\text{start}}\|_2^2 \big| \mathcal{F}_{\text{start}}\big]$$

$$\leq \mathbb{E}\big[F(\mathbf{x}_{\text{start}_u}) + c_u^{(s)}\|\mathbf{x}_{\text{start}_u} - \mathbf{x}_{\text{start}}\|_2^2 \big| \mathcal{F}_{\text{start}}\big] + \frac{2}{M}\mathbb{E}\big[\|\nabla F(\mathbf{x}_{\text{start}}) - \mathbf{v}_{\text{start}}\|_2^2 \big| \mathcal{F}_{\text{start}}\big](\text{end}_u - \text{start}_u). \tag{E.14}$$

We now try to telescope the above inequality. We first suppose that $u^* = \max\{0 \leq u < T_s : \text{start}_u < T\}$. Next we telescope (E.14) for $u = 0$ to $u^*$. Since we have $\text{start}_u = \text{end}_{u-1}, \text{start}_0 = \text{start}$ for $0 \leq u \leq u^*$, then we get

$$\mathbb{E}\bigg[\sum_{u=0}^{u^*}\sum_{j=\text{start}_u}^{\text{end}_u-1} \frac{\|\nabla F(\mathbf{x}_j)\|_2^2}{100M} + F(\mathbf{x}_{\text{end}_{u^*}}) + c_{u^*}^{(s)} \cdot \|\mathbf{x}_{\text{end}_{u^*}} - \mathbf{x}_{\text{start}}\|_2^2 \bigg| \mathcal{F}_{\text{start}}\bigg]$$

$$\leq F(\mathbf{x}_{\text{start}}) + \frac{2T_s}{M} \cdot \mathbb{E}\big[\|\nabla F(\mathbf{x}_{\text{start}}) - \mathbf{v}_{\text{start}}\|_2^2 \big| \mathcal{F}_{\text{start}}\big] \cdot \sum_{u=0}^{u^*}(\text{end}_u - \text{start}_u).$$

Since for $0 \leq u \leq u^*$, we have $\text{start}_u = \text{end}_{u-1}, \text{start}_0 = \text{start}, \text{end}_{u^*} = \text{end}$, and $c_{u^*}^{(s)} > c_{T_s}^{(s)}$, thus we have that

$$\mathbb{E}\bigg[\sum_{j=\text{start}}^{\text{end}-1} \frac{\|\nabla F(\mathbf{x}_j)\|_2^2}{100M} + F(\mathbf{x}_{\text{end}}) + c_{T_s}^{(s)} \cdot \|\mathbf{x}_{\text{end}} - \mathbf{x}_{\text{start}}\|_2^2 \bigg| \mathcal{F}_{\text{start}}\bigg]$$

$$\leq F(\mathbf{x}_{\text{start}}) + \frac{2}{M} \cdot \mathbb{E}\big[\|\nabla F(\mathbf{x}_{\text{start}}) - \mathbf{v}_{\text{start}}\|_2^2 \big| \mathcal{F}_{\text{start}}\big] \cdot (\text{end} - \text{start}).$$

Therefore, we have proved that Lemma D.3 still holds for $s$ and $p$. Then by mathematical induction, we have for all $1 \leq s \leq K$ and $p$ which satisfy $q\prod_{j=1}^{K} T_j \leq p \cdot \prod_{j=s}^{K} T_j < (p+1) \cdot \prod_{j=s}^{K} T_j \leq (q+1)\prod_{j=1}^{K} T_j$, Lemma D.3 holds. $\square$

# F Proof of Auxiliary Lemmas

Finally, we present the proof of Lemma E.2, which is a special case of Lemma D.3 when we choose $s = K$.

*Proof of Lemma E.2.* To simplify notations, we use $\mathbb{E}[\cdot]$ to denote the conditional expectation $\mathbb{E}[\cdot | \mathcal{F}_{p \cdot T_K}]$ in the rest of this proof. For $pT_K \leq pT_K + j < \min\{(p+1)T_K, T\}$, we denote $\mathbf{h}_{p \cdot T_K + j} = -(10M)^{-1} \cdot \mathbf{v}_{p \cdot T_K + j}$. According to the update in Algorithm 1 (Line 8), we have

$$\mathbf{x}_{p \cdot T_K + j + 1} = \mathbf{x}_{p \cdot T_K + j} + \mathbf{h}_{p \cdot T_K + j}, \tag{F.1}$$

which immediately implies

$$\begin{aligned}
F(\mathbf{x}_{p \cdot T_K + j + 1}) &= F(\mathbf{x}_{p \cdot T_K + j} + \mathbf{h}_{p \cdot T_K + j}) \\
&\leq F(\mathbf{x}_{p \cdot T_K + j}) + \langle \nabla F(\mathbf{x}_{p \cdot T_K + j}), \mathbf{h}_{p \cdot T_K + j} \rangle + \frac{L}{2} \|\mathbf{h}_{p \cdot T_K + j}\|_2^2 \tag{F.2} \\
&= \left[ \langle \mathbf{v}_{p \cdot T_K + j}, \mathbf{h}_{p \cdot T_K + j} \rangle + 5M \|\mathbf{h}_{p \cdot T_K + j}\|_2^2 \right] + F(\mathbf{x}_{p \cdot T_K + j}) \\
&\quad + \langle \nabla F(\mathbf{x}_{p \cdot T_K + j}) - \mathbf{v}_{p \cdot T_K + j}, \mathbf{h}_{p \cdot T_K + j} \rangle + \left( \frac{L}{2} - 5M \right) \|\mathbf{h}_{p \cdot T_K + j}\|_2^2 \\
&\leq F(\mathbf{x}_{p \cdot T_K + j}) + \langle \nabla F(\mathbf{x}_{p \cdot T_K + j}) - \mathbf{v}_{p \cdot T_K + j}, \mathbf{h}_{p \cdot T_K + j} \rangle + (L - 5M) \|\mathbf{h}_{p \cdot T_K + j}\|_2^2, \tag{F.3}
\end{aligned}$$

where (F.2) is due to the $L$-smoothness of $F$ and (F.3) holds because $\langle \mathbf{v}_{p \cdot T_K + j}, \mathbf{h}_{p \cdot T_K + j} \rangle + 5M \|\mathbf{h}_{p \cdot T_K + j}\|_2^2 = -5M \|\mathbf{h}_{p \cdot T_K + j}\|_2^2 \leq 0$. Further by Young's inequality, we obtain

$$\begin{aligned}
F(\mathbf{x}_{p \cdot T_K + j + 1}) &\leq F(\mathbf{x}_{p \cdot T_K + j}) + \frac{1}{2M} \|\nabla F(\mathbf{x}_{p \cdot T_K + j}) - \mathbf{v}_{p \cdot T_K + j}\|_2^2 + \left( \frac{M}{2} + L - 5M \right) \|\mathbf{h}_{p \cdot T_K + j}\|_2^2 \\
&\leq F(\mathbf{x}_{p \cdot T_K + j}) + \frac{1}{M} \|\nabla F(\mathbf{x}_{p \cdot T_K + j}) - \mathbf{v}_{p \cdot T_K + j}\|_2^2 - 3M \|\mathbf{h}_{p \cdot T_K + j}\|_2^2, \tag{F.4}
\end{aligned}$$

where the second inequality holds because $M > L$. Now we bound the term $c_{j+1}^{(K)} \|\mathbf{x}_{p \cdot T_K + j + 1} - \mathbf{x}_{p \cdot T_K}\|_2^2$. By (F.1) we have

$$\begin{aligned}
c_{j+1}^{(K)} \|\mathbf{x}_{p \cdot T_K + j + 1} - \mathbf{x}_{p \cdot T_K}\|_2^2 &= c_{j+1}^{(K)} \|\mathbf{x}_{p \cdot T_K + j} - \mathbf{x}_{p \cdot T_K} + \mathbf{h}_{p \cdot T_K + j}\|_2^2 \\
&= c_{j+1}^{(K)} \left[ \|\mathbf{x}_{p \cdot T_K + j} - \mathbf{x}_{p \cdot T_K}\|_2^2 + \|\mathbf{h}_{p \cdot T_K + j}\|_2^2 + 2 \langle \mathbf{x}_{p \cdot T_K + j} - \mathbf{x}_{p \cdot T_K}, \mathbf{h}_{p \cdot T_K + j} \rangle \right].
\end{aligned}$$

Applying Young's inequality yields

$$\begin{aligned}
c_{j+1}^{(K)} \|\mathbf{x}_{p \cdot T_K + j + 1} - \mathbf{x}_{p \cdot T_K}\|_2^2 &\leq c_{j+1}^{(K)} \bigg[ \|\mathbf{x}_{p \cdot T_K + j} - \mathbf{x}_{p \cdot T_K}\|_2^2 + \|\mathbf{h}_{p \cdot T_K + j}\|_2^2 \\
&\quad + \frac{1}{T_K} \|\mathbf{x}_{p \cdot T_K + j} - \mathbf{x}_{p \cdot T_K}\|_2^2 + T_K \|\mathbf{h}_{p \cdot T_K + j}\|_2^2 \bigg] \\
&= c_{j+1}^{(K)} \left[ \left( 1 + \frac{1}{T_K} \right) \|\mathbf{x}_{p \cdot T_K + j} - \mathbf{x}_{p \cdot T_K}\|_2^2 + (1 + T_K) \|\mathbf{h}_{p \cdot T_K + j}\|_2^2 \right], \tag{F.5}
\end{aligned}$$

Adding up inequalities (F.5) and (F.4), we get

$$F(\mathbf{x}_{p \cdot T_K + j + 1}) + c_{j+1}^{(K)} \|\mathbf{x}_{p \cdot T_K + j + 1} - \mathbf{x}_{p \cdot T_K}\|_2^2$$

$$\leq F(\mathbf{x}_{p\cdot T_K+j}) + \frac{1}{M}\|\nabla F(\mathbf{x}_{p\cdot T_K+j}) - \mathbf{v}_{p\cdot T_K+j}\|_2^2 - [3M - c_{j+1}^{(K)}(1+T_K)]\|\mathbf{h}_{p\cdot T_K+j}\|_2^2$$
$$+ c_{j+1}^{(K)}\left(1 + \frac{1}{T_K}\right)\|\mathbf{x}_{p\cdot T_K+j} - \mathbf{x}_{p\cdot T_K}\|_2^2$$
$$\leq F(\mathbf{x}_{p\cdot T_K+j}) + \frac{1}{M}\|\nabla F(\mathbf{x}_{p\cdot T_K+j}) - \mathbf{v}_{p\cdot T_K+j}\|_2^2 - 2M\|\mathbf{h}_{p\cdot T_K+j}\|_2^2$$
$$+ c_{j+1}^{(K)}\left(1 + \frac{1}{T_K}\right)\|\mathbf{x}_{p\cdot T_K+j} - \mathbf{x}_{p\cdot T_K}\|_2^2, \tag{F.6}$$

where the last inequality holds due to the fact that $c_{j+1}^{(K)}(1+T_K) < M$ by Lemma D.2. Next we bound $\|\nabla F(\mathbf{x}_{p\cdot T_K+j})\|_2^2$ with $\|\mathbf{h}_{p\cdot T_K+j}\|_2^2$. Note that by (F.1)

$$\|\nabla F(\mathbf{x}_{p\cdot T_K+j})\|_2^2 = \left\|[\nabla F(\mathbf{x}_{p\cdot T_K+j}) - \mathbf{v}_{p\cdot T_K+j}] - 10M\mathbf{h}_{p\cdot T_K+j}\right\|_2^2$$
$$\leq 2\big(\|\nabla F(\mathbf{x}_{p\cdot T_K+j}) - \mathbf{v}_{p\cdot T_K+j}\|_2^2 + 100M^2\|\mathbf{h}_{p\cdot T_K+j}\|_2^2\big),$$

which immediately implies

$$-2M\|\mathbf{h}_{p\cdot T_K+j}\|_2^2 \leq \frac{2}{100M}\big(\|\nabla F(\mathbf{x}_{p\cdot T_K+j}) - \mathbf{v}_{p\cdot T_K+j}\|_2^2 - \frac{1}{100M}\|\nabla F(\mathbf{x}_{p\cdot T_K+j})\|_2^2. \tag{F.7}$$

Plugging (F.7) into (F.6), we have

$$F(\mathbf{x}_{p\cdot T_K+j+1}) + c_{j+1}^{(K)}\|\mathbf{x}_{p\cdot T_K+j+1} - \mathbf{x}_{p\cdot T_K}\|_2^2$$
$$\leq F(\mathbf{x}_{p\cdot T_K+j}) + \frac{1}{M}\|\nabla F(\mathbf{x}_{p\cdot T_K+j}) - \mathbf{v}_{p\cdot T_K+j}\|_2^2 + \frac{1}{50M}\cdot\|\nabla F(\mathbf{x}_{p\cdot T_K+j}) - \mathbf{v}_{p\cdot T_K+j}\|_2^2$$
$$- \frac{1}{100M}\|\nabla F(\mathbf{x}_{p\cdot T_K+j})\|_2^2 + c_{j+1}^{(K)}\left(1 + \frac{1}{T_K}\right)\|\mathbf{x}_{p\cdot T_K+j} - \mathbf{x}_{p\cdot T_K}\|_2^2$$
$$\leq F(\mathbf{x}_{p\cdot T_K+j}) + \frac{2}{M}\|\nabla F(\mathbf{x}_{p\cdot T_K+j}) - \mathbf{v}_{p\cdot T_K+j}\|_2^2 - \frac{1}{100M}\|\nabla F(\mathbf{x}_{p\cdot T_K+j})\|_2^2$$
$$+ c_{j+1}^{(K)}\left(1 + \frac{1}{T_K}\right)\|\mathbf{x}_{p\cdot T_K+j} - \mathbf{x}_{p\cdot T_K}\|_2^2. \tag{F.8}$$

Next we bound $\|\nabla F(\mathbf{x}_{p\cdot T_K+j}) - \mathbf{v}_{p\cdot T_K+j}\|_2^2$. First, by Lemma E.3 we have

$$\mathbb{E}\left\|\nabla F(\mathbf{x}_{p\cdot T_K+j}^{(K)}) - \mathbf{v}_{p\cdot T_K+j}^{(K)}\right\|_2^2 \leq \frac{L^2}{B_K}\mathbb{E}\left\|\mathbf{x}_{p\cdot T_K+j}^{(K)} - \mathbf{x}_{p\cdot T_K+j}^{(K-1)}\right\|_2^2 + \mathbb{E}\left\|\nabla F(\mathbf{x}_{p\cdot T_K+j}^{(K-1)}) - \mathbf{v}_{p\cdot T_K+j}^{(K-1)}\right\|_2^2.$$

Since $\mathbf{x}_{p\cdot T_K+j}^{(K)} = \mathbf{x}_{p\cdot T_K+j}, \mathbf{v}_{p\cdot T_K+j}^{(K)} = \mathbf{v}_{p\cdot T_K+j}, \mathbf{x}_{p\cdot T_K+j}^{(K-1)} = \mathbf{x}_{p\cdot T_K}$ and $\mathbf{v}_{p\cdot T_K+j}^{(K-1)} = \mathbf{v}_{p\cdot T_K}$, we have

$$\mathbb{E}\|\nabla F(\mathbf{x}_{p\cdot T_K+j}) - \mathbf{v}_{p\cdot T_K+j}\|_2^2 \leq \frac{L^2}{B_K}\mathbb{E}\|\mathbf{x}_{p\cdot T_K+j} - \mathbf{x}_{p\cdot T_K}\|_2^2 + \mathbb{E}\|\nabla F(\mathbf{x}_{p\cdot T_K}) - \mathbf{v}_{p\cdot T_K}\|_2^2. \tag{F.9}$$

Taking expectation $\mathbb{E}[\cdot]$ with (F.8) and plugging (F.9) into (F.8), we obtain

$$\mathbb{E}\bigg[F(\mathbf{x}_{p\cdot T_K+j+1}) + c_{j+1}^{(K)}\|\mathbf{x}_{p\cdot T_K+j+1} - \mathbf{x}_{p\cdot T_K}\|_2^2 + \frac{1}{100M}\|\nabla F(\mathbf{x}_{p\cdot T_K+j})\|_2^2\bigg]$$
$$\leq \mathbb{E}\bigg[F(\mathbf{x}_{p\cdot T_K+j}) + \left(c_{j+1}^{(K)}\left(1+\frac{1}{T_K}\right) + \frac{3L^2}{B_KM}\right)\|\mathbf{x}_{p\cdot T_K+j} - \mathbf{x}_{p\cdot T_K}\|_2^2 + \frac{2}{M}\|\nabla F(\mathbf{x}_{p\cdot T_K}) - \mathbf{v}_{p\cdot T_K}\|_2^2\bigg]$$

$$= \mathbb{E}\bigg[F(\mathbf{x}_{p\cdot T_K+j}) + c_j^{(K)}\|\mathbf{x}_{p\cdot T_K+j} - \mathbf{x}_{p\cdot T_K}\|_2^2 + \frac{2}{M}\cdot \|\nabla F(\mathbf{x}_{p\cdot T_K}) - \mathbf{v}_{p\cdot T_K}\|_2^2\bigg], \tag{F.10}$$

where (F.10) holds because we have $c_j^{(K)} = c_{j+1}^{(K)}(1 + 1/T_K) + 3L^2/(B_K M)$ by Definition D.1. Telescoping (F.10) for $j = 0$ to $\text{end} - \text{start} - 1$, we have

$$\mathbb{E}\big[F(\mathbf{x}_{\text{end}}) + c_{T_K}^{(K)}\cdot \|\mathbf{x}_{\text{end}} - \mathbf{x}_{\text{start}}\|_2^2\big] + \frac{1}{100M}\sum_{j=\text{start}}^{\text{end}-1}\mathbb{E}\|\nabla F(\mathbf{x}_j)\|_2^2$$

$$\leq \mathbb{E}\big[F(\mathbf{x}_{\text{end}}) + c_{\text{end}-\text{start}}^{(K)}\cdot \|\mathbf{x}_{\text{end}} - \mathbf{x}_{\text{start}}\|_2^2\big] + \frac{1}{100M}\sum_{j=\text{start}}^{\text{end}-1}\mathbb{E}\|\nabla F(\mathbf{x}_j)\|_2^2$$

$$\leq F(\mathbf{x}_{\text{start}}) + \frac{2(\text{end}-\text{start})}{M}\cdot \mathbb{E}\|\nabla F(\mathbf{x}_{\text{start}}) - \mathbf{v}_{\text{start}}\|_2^2,$$

which completes the proof. □